\documentclass[10pt,twocolumn,letterpaper]{article}

\usepackage{iccv}
\usepackage{times}
\usepackage{epsfig}
\usepackage{amsmath}
\usepackage{amssymb}
\usepackage{graphicx}
\usepackage{subfig}
\usepackage{stackengine}
\usepackage{color}
\usepackage{CJK}

\usepackage[svgpath=./fig/]{svg}

\newcommand{\figsubtitle}[2]{%
    \begin{minipage}{#2} % width
        \begin{flushleft}%
            \hrulefill\\
            {\footnotesize \textsf{#1}}% content
        \end{flushleft}%
    \end{minipage}
}

\iccvfinalcopy % *** Uncomment this line for the final submission

\def\iccvPaperID{6433} % *** Enter the ICCV Paper ID here

\ificcvfinal\fi
\begin{document}

%%%%%%%%% TITLE
\title{A Learned Representation for Scalable Vector Graphics}
% Alternative Title (feel free to change):
%  A Learned Representation of Scalable Vector Graphics

\author{Raphael Gontijo Lopes\thanks{Work done as a member of the Google AI Residency Program \url{(g.co/airesidency)}}, David Ha, Douglas Eck, Jonathon Shlens\\
Google Brain\\
{\tt\small \{iraphael, hadavid, deck, shlens\}@google.com}}

\newcommand{\classvalue}[1]{\texttt{#1}}
\newcommand{\svg}[1]{\texttt{#1}}

\maketitle

\begin{abstract}
Dramatic advances in generative models have resulted in near photographic quality for artificially rendered faces, animals and other objects in the natural world.
In spite of such advances, a higher level understanding of vision and imagery does not arise from exhaustively modeling an object, but instead identifying higher-level attributes that best summarize the aspects of an object. In this work we attempt to model the drawing process of fonts by building sequential generative models of vector graphics. This model has the benefit of providing a scale-invariant representation for imagery whose latent representation may be systematically manipulated and exploited to perform style propagation. We demonstrate these results on a large dataset of fonts and highlight how such a model captures the statistical dependencies and richness of this dataset. We envision that our model can find use as a tool for graphic designers to facilitate font design.
\end{abstract}
\section{Introduction}
\label{sec:intro}

% WIP:
\begin{figure}[t]
    \centering%
    \subfloat{%
        \figsubtitle{Learned Vector Graphics Representation}{0.33\textwidth}%
        \figsubtitle{Pixel Counterpart}{0.13\textwidth}%
    }%
    \hfill%
    \subfloat{%
        \includegraphics[trim={0.4in 0.4in 0.4in 0.4in},clip=true,draft=false,width=0.465\textwidth]{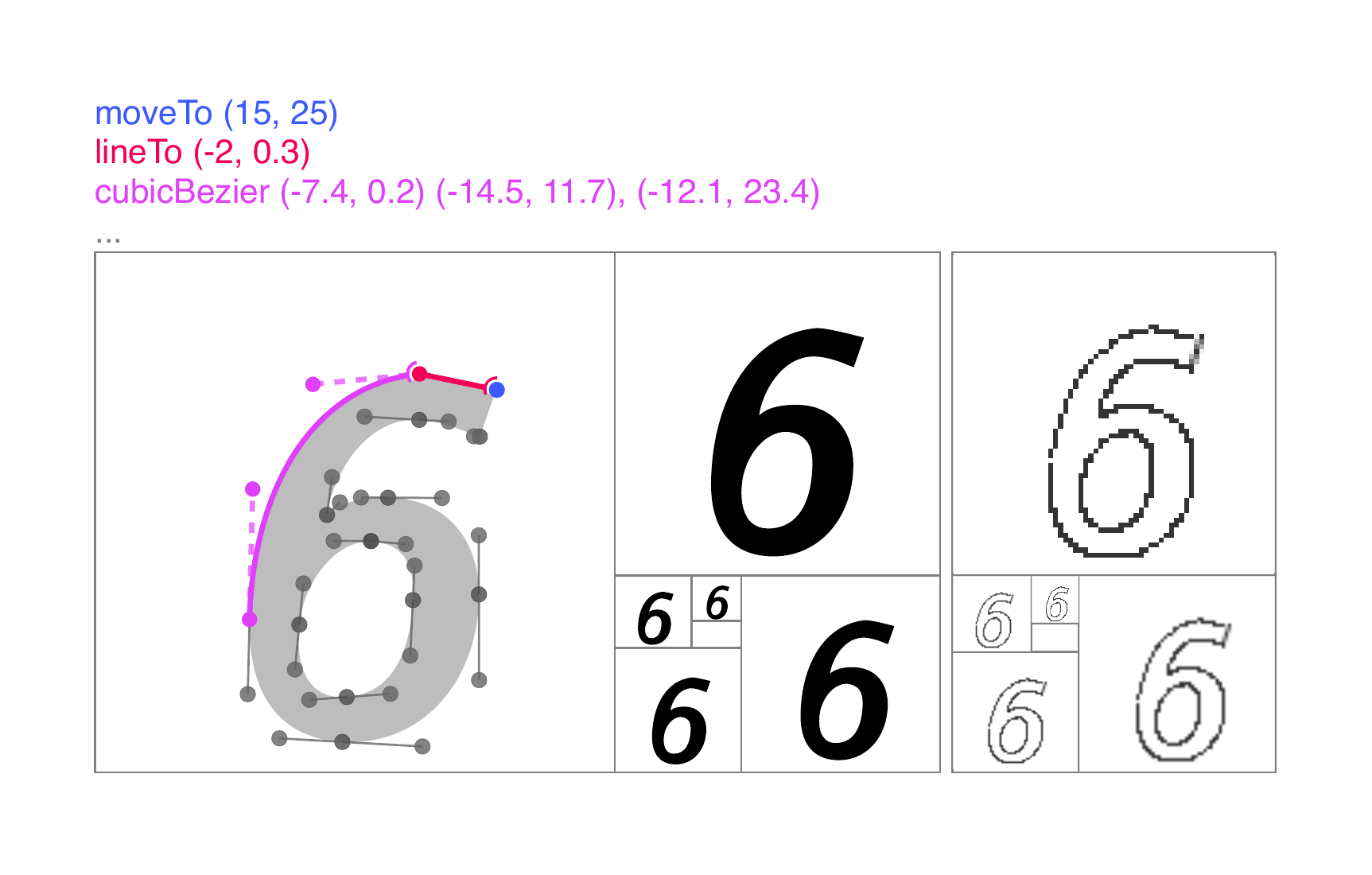}%
    }%
    \hfill%
    \subfloat{%
        \figsubtitle{Conveying Different Styles}{0.46\textwidth}%
    }%
    \hfill%
    \vspace{-0.011\textwidth}%
    \subfloat{%
        \centering
        \includegraphics[trim={0 0 0 0.1175\textwidth},clip=true,draft=false,width=0.3995\textwidth]{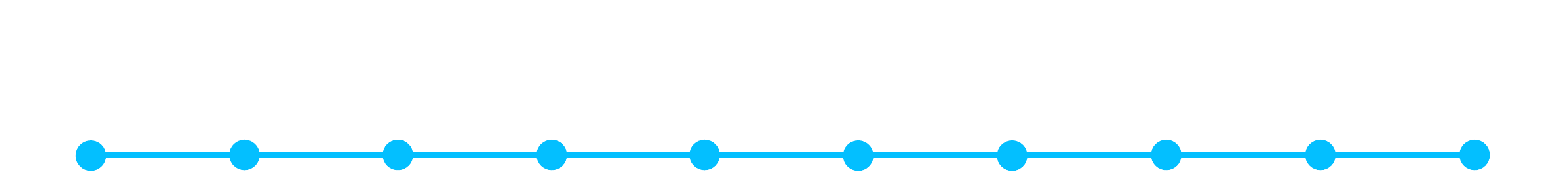}
    }%
    \vspace{-0.011\textwidth}%
    \hfill%
    \subfloat{%   
        \includegraphics[draft=false,width=0.3995\textwidth]{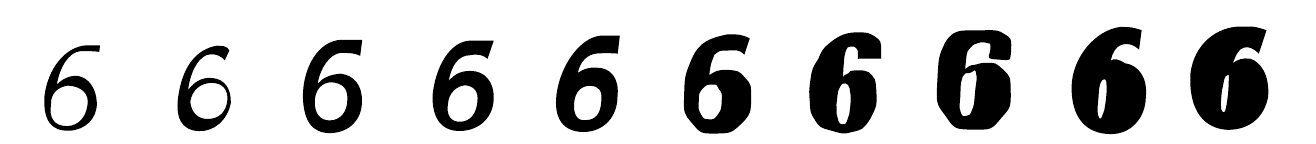}%
    }
    \caption{\textbf{Learning fonts in a native command space}. Unlike pixels, scalable vector graphics (SVG) \cite{ferraiolo2000scalable} are scale-invariant representations whose parameterizations may be systematically adjusted to convey different styles. All vector images are samples from a generative model of the SVG specification.}   
    % \caption{\textbf{Learning fonts in a native command space}. Unlike pixels (right), scalable vector graphics (SVG) \cite{ferraiolo2000scalable} are scale-invariant representations whose parameterizations (top) may be systematically adjusted to convey different styles (bottom). All vector images are samples from a generative model of the SVG specification.}   
    \label{fig:svg}
\end{figure}

The last few years have witnessed dramatic advances in generative models of images that produce near photographic quality imagery of human faces, animals, and natural objects \cite{brock2018large, karras2018style, kingma2018glow}.
These models provide an exhaustive characterization of natural image statistics \cite{simoncelli2001natural} and represent a significant advance in this domain. However, these advances in image synthesis ignore an important facet of how humans interpret raw visual information \cite{reisberg2010cognition}, 
namely that humans seem to exploit \textit{structured} representations of visual concepts \cite{lake2017building, hofstadter1995fluid}.
Structured representations may be readily employed to aid generalization and efficient learning by identifying higher level primitives for conveying visual information \cite{lake2015human} or provide building blocks for creative exploration \cite{hofstadter1995fluid, hofstadter1993letter}. This may be best seen in human drawing, where techniques such as \textit{gesture drawing} \cite{picasso_1945} emphasize parsimony for capturing higher level semantics and actions with minimal graphical content \cite{stanchfield2007gesture}.

%deck@: I like this footnote. But we might need the rights %to picasso bulls in order to use it.
%\footnote{See, for instance, ``Picasso's bulls'' as a dramatic example.}. 

Our goal is to train a drawing model by presenting it with a large set of example images \cite{ha2017neural, ganin2018synthesizing}. To succeed, the model needs to learn both the underlying structure
%(i.e. "disentangled variance") 
in those images and to generate drawings based on the learned representation \cite{bengio2013overview}.
In computer vision this is referred to as an ``inverse graphics" problem \cite{mansinghka2013approximate, kulkarni2015deep, jampani2015informed, nair2008analysis}. % deck@
% This jump to discrete symbols is a bit odd because 
% we haven't yet talked about non-pixel-based rendering
% Just one sentence? 
In our case the output representation is not pixels but rather a sequence of discrete instructions for rendering a drawing on a graphics engine. This poses dual challenges in terms of learning discrete representations for latent variables \cite{van2017neural, jang2016categorical, maddison2016concrete} and performing optimization through a non-differential graphics engine (but see \cite{loper2014opendr, kulkarni2015deep}). %Tackling the structured learning problem for visual imagery is challenging.
Previous approaches focused on program synthesis approaches \cite{lake2015human, ellis2018learning} or employing reinforcement and adversarial learning \cite{ganin2018synthesizing}. We instead focus on a subset of this domain where we think we can make progress and improves the generality of the approach.

Font generation represents a 30 year old problem posited as a constrained but diverse domain for understanding higher level perception and creativity \cite{hofstadter1995fluid}. Early research attempted to heuristically systematize the creation of fonts for expressing the identity of characters (e.g. \classvalue{a}, \classvalue{2}) as well as stylistic elements constituting the ``spirit'' of a font \cite{hofstadter1993letter}. Although such work provides 
% I'd avoid "brittle" as a bit pejorative (deck@)
% even if Jon Rehling maybe said it :)
great inspiration, the results were limited by a reliance on heuristics and a lack of a learned, structured representation \cite{ rehling2001letter}. Subsequent work for learning representations for fonts focused on models with simple parameterizations \cite{lau2009learning}, template matching \cite{suveeranont2010example}, example-based hints \cite{zongker2000example}, or more recently, learning manifolds for detailed geometric annotations \cite{campbell2014learning}.

We instead focus the problem on generating fonts specified with Scalable Vector Graphics (SVG) -- a common file format for fonts, human drawings, designs and illustrations \cite{ferraiolo2000scalable}. SVG's are a compact, scale-invariant representation that may be rendered on most web browsers. SVG's specify an illustration as a sequence of a higher-level commands paired with numerical arguments (Figure \ref{fig:svg}, top). We take inspiration from the literature on generative models of images in rasterized pixel space \cite{graves2013generating, van2016conditional}. Such models provide powerful auto-regressive formulations for discrete, sequential data \cite{graves2013generating, van2016conditional} and may be applied to rasterized renderings of drawings \cite{ha2017neural}. We extend these approaches to the generation of sequences of SVG commands for the inference of individual font characters. The goal of this work is to build a tool to learn a representation for font characters and style that may be extended to other artistic domains \cite{clouatre2019figr,sketchy2016, ha2017neural}, or exploited as an intelligent assistant for font creation \cite{carter2017using}. In this work, our main contributions are as follows:

\begin{itemize}
    \item Build a generative model for scalable vector graphics (SVG) images and apply this to a large-scale dataset of 14\,M font characters. %\todo{rapha: 14M glyphs! not fonts. but should we use the "glyph" jargon here without having introduced it?}
    \item Demonstrate that the generative model provides a perceptually smooth latent representation for font styles that captures a large amount of diversity and is consistent across individual characters.
    \item Exploit the latent representation from the model to infer complete SVG fontsets from a single (or multiple) characters of a font.
    \item Identify semantically meaningful directions in the latent representation to globally manipulate font style.
    %\item Highlight the limitations of this modeling approach for generating high quality fonts.
\end{itemize}

\section{Related Work}

\subsection{Generative models of images}

Generative models of images have generally followed two distinct directions. %hadavid: removed: The first strategy has abandoned the goal of building a probabilistic model and instead focused on developing a loss function for images \cite{goodfellow2014generative}. 
Generative adversarial networks \cite{goodfellow2014generative} have demonstrated impressive advances \cite{radford2015unsupervised, goodfellow2014generative} over the last few years resulting in models that generate high resolution imagery nearly indistinguishable from real photographs \cite{karras2018style, brock2018large}.

A second direction has pursued building probabilistic models largely focused on invertible representations \cite{dinh2016density, kingma2018glow}. Such models are highly tractable and do not suffer from training instabilities largely attributable to saddle-point optimization \cite{goodfellow2014generative}. Additionally, such models afford a true probabilistic model in which the quality of the model may be measured with well-characterized objectives such as log-likelihood.

\subsection{Autoregressive generative models}

One method for vastly improving the quality of generative models with unsupervised objectives is to break the problem of joint prediction into a conditional, sequential prediction task. Each step of the conditional prediction task may be expressed with a sequential model (e.g. \cite{hochreiter1997long}) trained in an autoregressive fashion. Such models are often trained with a teacher-forcing training strategy, but more sophisticated methods may be employed \cite{bengio2015scheduled}.

Autoregressive models have demonstrated great success in speech synthesis \cite{oord2016wavenet} and unsupervised learning tasks \cite{van2017neural} across multiple domains. Variants of autoregressive models paired with more sophisticated density modeling \cite{bishop1994mixture} have been employed for sequentially generating handwriting \cite{graves2013generating}.

\subsection{Modeling higher level languages}

The task of learning an algorithm from examples has been widely studied. Lines of work vary from directly modeling computation \cite{kaiser2015neural} to learning a hierarchical composition of given computational primitives \cite{fox2018parametrized}. Of particular relevance are efforts that learn to infer graphics programs from the visual features they render, often using constructs like variable bindings, loops, or simple conditionals \cite{ellis2018learning}.

The most comparable methods to this work yield impressive results on unsupervised induction of programs usable by a given graphics engine \cite{ganin2018synthesizing}. As their setup is non differentiable, they use the REINFORCE \cite{williams1992simple} algorithm to perform adversarial training \cite{goodfellow2014generative}. This method achieves impressive results despite not relying on labelled paired data. However, it tends to draw over previously-generated drawings, especially later in the generation process. While this could be suitable for modelling the generation of a 32x32 rastered image, SVGs require a certain level of precision in order to scale with few perceptible issues.

\subsection{Learning representations for fonts}
% zongker2000example - find correspondences between points in chars of diff fonts to transfer typehinting between them
% ^not mentioning this anymore

% suveeranont2010example - modelling of font style by example. can do style propagation for automatic font generation, but does it by introducing simplifying assumptions (eg. all chars in a class have a common skeleton)

% lau2009learning - according to learning a manifold of fonts, this work proposes to learn a parametric model with constraints from examples. However, the parametric model is simple and allows the user to only adjust parameters like its weight or width.

% rehling2001letter - LetterSpirit system that automatically identifies class and style from human provided examples, and allows style propagation, but makes huge simplifying assumptions like all chars being composed of very specific set of discrete lines.

% hofstadter1995fluid - describes general concepts of learning by analogy, but also includes a chapter on LetterSpirit

% hofstadter1993letter - LetterSpirit

Previous work has focused on enabling propagation of style between classes by identifying the class and style from high level features of a single character \cite{rehling2001letter, hofstadter1995fluid, hofstadter1993letter} or by finding correspondences between these features of different characters \cite{suveeranont2010example}. These features are often simplifications, such as character skeletons, which diminish the flexibility of the methods. Other work directly tackles style manipulation of generated full characters \cite{lau2009learning}, but uses a simple parametric model that allows the user to only adjust parameters like its weight or width.

% Previous work has focused on finding correspondences between discrete high-level features of characters to enable propagation of style \cite{suveeranont2010example}.
% They do it by making simplifying assumptions about which high level features to model (e.g.: character skeleton, specific points in similar fonts). This is the same issue that learning a manifold of fonts has.
% Other work automatically identifies class and style from a single character and allows style propagation to other classes \cite{rehling2001letter, hofstadter1995fluid, hofstadter1993letter}. However, all of these makes simplifying assumptions about which features to model, such as composing characters of only a small set of discrete lines.

% Other work directly tackles style manipulation of generated full characters \cite{lau2009learning}, but uses a simple parametric model that allows the user to only adjust parameters like its weight or width.

The most relevant works are those that attempt to learn a manifold of font style. Some unpublished work has explored how probabilistic methods may model pixel-based representations of font style \cite{loh2018space}. The model learned semantically-meaningful latent spaces which can manipulate rasterized font images.
%The application of this work is limited the fact that the model operates in pixel-space for representing fonts.
More directly comparable, recent work learned energy models to capture the relationship between discretized points along the outlines of each character in order to address font generation and extrapolation \cite{campbell2014learning}. 
This method yields impressive results for extrapolating between very few examples, but is limited by the need of having all characters of a certain class be composed of equivalent shapes
%, which also comes at the cost of modelling many varieties of font styles.
Additionally, the model discretely approximates points on a character's outline which may lead to visual artifacts at larger spatial scales.

\section{Methods}

\begin{figure}[t!]
    \begin{center}
    \figsubtitle{Image Autoencoder}{0.24\textwidth}%
    \hspace{0.02\textwidth}%
    \figsubtitle{SVG Decoder}{0.15\textwidth}%
    \vspace{0.01\textwidth}\\
    \includegraphics[draft=false,height=0.1\textwidth]{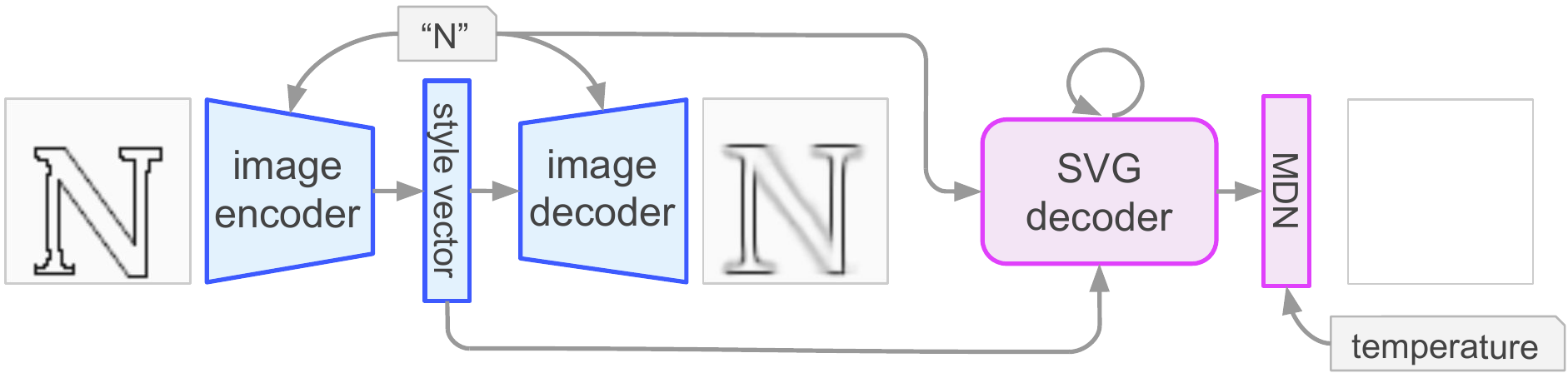}
    \llap{\raisebox{0.15in}{\hspace{-0.05in}\mbox{
        \includegraphics[draft=false,height=0.05
    \textwidth]{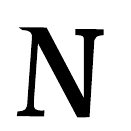}
    }\hspace{0.05in}}}
    \end{center}
    \caption{\textbf{Model architecture}. Visual similarity between SVGs is learned by a class-conditioned, convolutional variational autoencoder (VAE) \cite{kingma2013auto, ha2017neural} on a rendered representation (blue). The class label and learned representation $z$ are provided as input to a model that decodes SVG commands (purple). The SVG decoder consists of stacked LSTM's \cite{hochreiter1997long} followed by a Mixture Density Network (MDN) \cite{graves2013generating, bishop1994mixture}. See text for details.}
    %\caption{\textbf{Model architecture}. We first learn a visual similarity metric between SVGs by training a convolutional variational autoencoder (purple) on a rendered representation of the icons. Both the encoder and decoder of this VAE is conditioned on the icon class (e.g.: 0-9, a-z, A-Z) and carefully tuned to ensure that it's z representation is a class-invariant style representation of the given icon. After training, this learned representation is given, along with the desired class, to an SVG decoder, parameterized as an LSTM (red) which outputs SVG commands. The SVG decoder is equipped with a Mixture Density Network (MDN) which allows it to model multimodal distributions, in similar fashion to sketch-rnn ((CITE)). The SVG commands outputted are composed of a probability distribution over command types (lineTo, moveTo, cubicBezierCurve, etc) concatenated with the command's arguments (e.g.: relative x, y positions for control points of the bezier curve). See text for more details. ((NOTE: probably dont go into too much detail here))}
    \label{fig:model}
\end{figure}

\subsection{Data}
% \todo{Is the name 'glyphs' just an internal name for the dataset? if we intend to release a subset of this data, perhaps we should use this name or select one? rapha says: it's also a technical term for "instance of a character". So it might be ok to come up with a derivative of the term to be the dataset name}
%\todo{there's way too much information here. we should move most of it to appendix}

% \todo{Can we release a fraction of this dataset? Probably not but perhaps a public subset exists somewhere. rapha says: perhaps a very small subset. But we can definitely find a subset. I've found at least 24 fonts with creative commons licenses in the dataset. Current work on this involves figuring out which licenses are fine, and then finding which fonts use those.}

We compiled a font dataset composed of $14$\,M examples across 62 characters (i.e. \classvalue{0}-\classvalue{9}, \classvalue{a}-\classvalue{z}, \classvalue{A}-\classvalue{Z}), which we term \texttt{SVG-Fonts}. The dataset consists of fonts in a common font format (SFD) \footnote{\texttt{https://fontforge.github.io}}, excluding examples where the unicode ID does not match the targeted 62 character set specified above. In spite of the filtering, label noise exists across the roughly 220\,K fonts examined. 

We employed a one-to-one mapping from SFD to SVG file format using a subset of 4 SVG commands: \svg{moveTo}, \svg{lineTo}, \svg{cubicBezier} and \svg{EOS}. SVG commands were normalized by starting from the top-most command and ordering in a clockwise direction. In preliminary experiments, we found it advantageous to specify command arguments using relative positioning information. See Appendix for more details of the dataset collection and normalization.

The final dataset consists of a sequence of commands specified in tuples. Each item in the sequence consists of a discrete selection of an SVG command paired with a set of normalized, floating-point numbers specifying command arguments. We restricted the dataset to only 4 SVG command types and examples with fewer then 50 commands to aid learning but these restrictions may be relaxed to represent the complete SVG language. In comparison, note that \cite{ganin2018synthesizing} restricted inference to 20 actions for generating images. Finally, we partitioned the dataset into $12.6$M and $1.4$M examples for training and testing
% ensuring that each glyph is only represented in a single dataset split
\footnote{We plan to open-source tools to reproduce the construction of a subset of the dataset, as well as code to train the proposed model upon acceptance to a peer-reviewed conference.}. % hadavid: modified the open-source to convey more clearly the fact that it is not our fault that we can't open source the dataset, but due to original font copyrights.

\subsection{Network Architecture}
%\todo{there might be too many details here, make it only a 1-2 paragraph overview of network architecture. move the details of the network architecture for the Appendix.}

The proposed model consists of a variational autoencoder (VAE) \cite{kingma2013auto, ha2017neural} and an autoregressive SVG decoder implemented in Tensor2Tensor \cite{vaswani2018tensor2tensor}. Figure \ref{fig:model} provides a diagram of the architecture but please see the Appendix for details. Briefly, the VAE consists of a convolutional encoder and decoder paired with instance normalization conditioned
on the label (e.g. \classvalue{a}, \classvalue{2}, etc.) \cite{dumoulin2017learned, perez2018film}. The VAE is trained as a class-conditioned autoencoder resulting in a latent code $z$ that is largely class-independent \cite{kingma2014semi}. In preliminary experiments, we found that 32 dimensions provides a reasonable balance between expressivity and tractability. Note that the latent code consists of $\mu$ and $\sigma$ - the mean and standard deviation of a multivariate Gaussian that may be sampled at test time.

The SVG decoder consists of 4 stacked LSTMs \cite{hochreiter1997long} trained with dropout \cite{srivastava2014dropout, zaremba2014recurrent,semeniuta2016recurrent}.  The final layer is a Mixture Density Network (MDN) \cite{bishop1994mixture, graves2013generating} that may be stochastically sampled at test time.
The LSTM receives as input the previous sampled MDN output, concatenated with the discrete class label and the latent style representation $z$. The SVG decoder's loss is composed of a softmax cross-entropy loss over one-hot SVG commands plus the MDN loss applied to the real-valued arguments.

%The encoded representation from the VAE $z$ representation of the input image, in the form of a mean $\mu$ and the log of the standard deviation $log(\sigma)$. At training time, we use the reparameterization trick (\todo{cite?}) to sample $z$ from these statistics, which is then fed to the image decoder.

%The SVG decoder consists of 4 stacked LSTMs cells, trained with feed-forward dropout \cite{srivastava2014dropout, zaremba2014recurrent}, as well as recurrent dropout (\todo{cite}). The decoder's topmost layer consists of a Mixture Density Network (MDN) (\todo{cite}). It's hidden state is initialized by conditioning on $z$. At each time-step, the LSTM receives as input the previous time-step's sampled MDN output, concatenated with the glyph's class and the $z$ representation.

%Our model is implemented using the Tensor2Tensor framework (\todo{cite}) and we plan on open sourcing it, along with a subset of the dataset.

%\subsection{Training and Evaluation}

In principle, the model may be trained end-to-end, but we found it simpler to train the two parts of the model separately. %\todo{Rapha, verify? rapha says: nothing is stopping us from being able to train the model end-to-end, but I assume that doing so might require a few more tricks (like scheduling an alpha on the SVG loss) to make sure that the visual loss dominates the bottleneck. So I'd say the statement above is true, just that there is nuance to it that we might not need to mention}.
The VAE is trained using pixel renderings of the fonts using the Adam optimizer ($\epsilon=10^{-6}$) \cite{kingma2014adam} for $3$ epochs. % The number of epochs is also an approximation, and has the nuance of involving warm-up and decay schedules, as well as picking examples with replacement (so we might not actually have seen each example 3 times at the end).
We employ a high value of $\beta$ \cite{higgins2017beta}, and tune the number of free bits using cross-validation \cite{kingma2016improved}. After convergence, the weights of the VAE are frozen, and the SVG decoder is trained to output the SVG commands from the latent representation $z$ using teacher forcing \cite{williams1989learning}.

Note that both the VAE and MDN are probabilistic models that may be sampled many times during evaluation. The results shown here are the selected best out of 10 samples. Please see Appendix for modeling, training and evaluation details.

%The SVG decoder's loss is composed of a softmax cross-entropy loss between the one-hot command type encoding, added to the MDN loss applied to the real-valued arguments (\todo{potentially describe this loss using math}).

%Because the MDN is probabilistic, it allows us to sample commands with multiple variances, by increasing or decreasing the temperature parameter (\todo{describe with math?}). This allows the model to account for information not available in $z$ with stochasticity. It also allows us to re-sample icons for evaluation.
\section{Results}

\begin{figure}[t]
    \begin{center}
    \includegraphics[draft=false,width=0.45\textwidth]{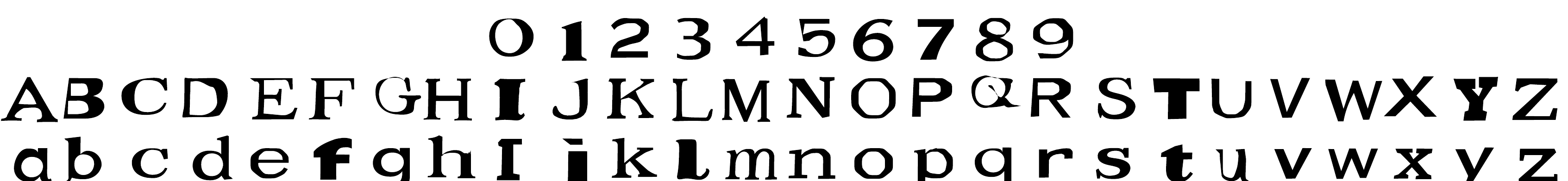}
    \includegraphics[draft=false,width=0.45\textwidth]{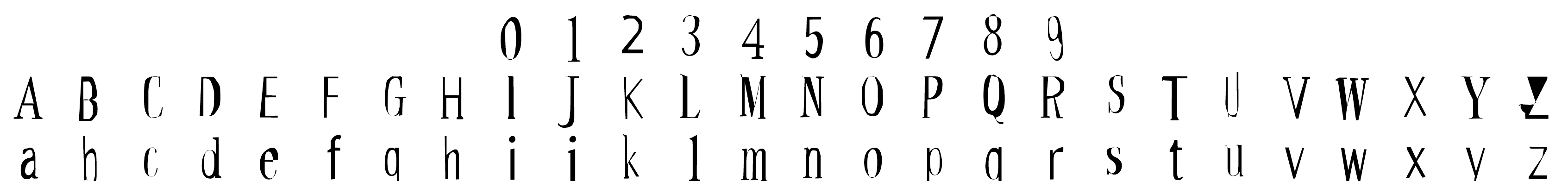}
    \includegraphics[draft=false,width=0.45\textwidth]{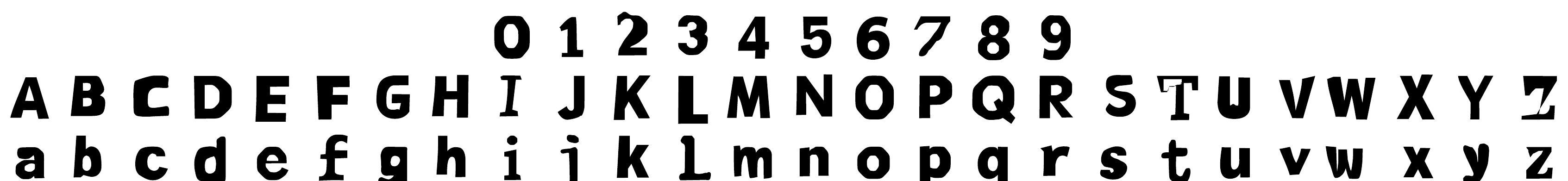}
    \includegraphics[draft=false,width=0.45\textwidth]{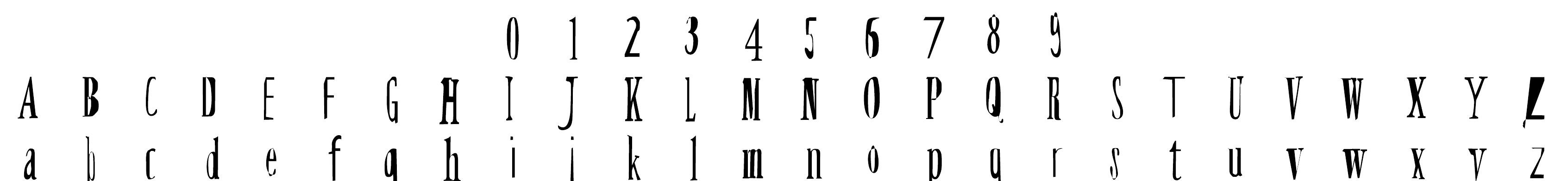}
    \includegraphics[draft=false,width=0.45\textwidth]{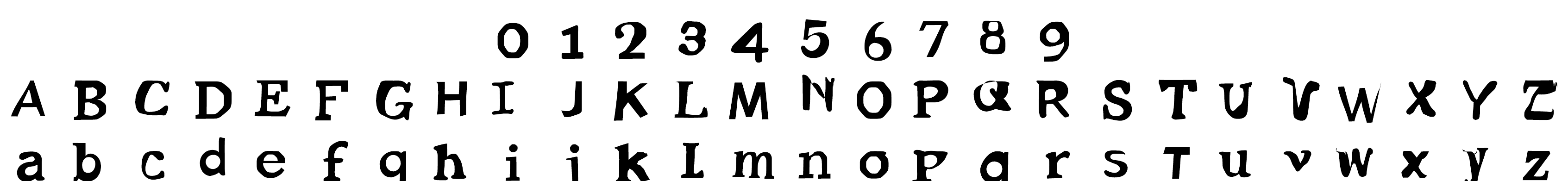}
    \end{center}
   \caption{\textbf{Selected examples of generated fonts}. Examples generated by sampling a random latent representation $z$ and running the SVG decoder by conditioning on $z$ and all class labels. Each font character is selected as the best of 10 samples. See Figures \ref{fig:more-random-fonts} and \ref{fig:bad-random-fonts} in Appendix for additional examples.}
    \label{fig:random-fonts}
\end{figure}
% rapha: made this figure small. We can revert if necessary but there is a similar large one in the appendix so I figured we could save space here.
% \begin{figure*}[t]
%     \begin{center}
%     \includegraphics[draft=false,height=1.35cm]{fig/full_alphabets/1.pdf}
%     \includegraphics[draft=false,height=1.35cm]{fig/full_alphabets/5.pdf}
%     \includegraphics[draft=false,height=1.35cm]{fig/full_alphabets/4.pdf}
%     \includegraphics[draft=false,height=1.35cm]{fig/full_alphabets/14.pdf}
%     \includegraphics[draft=false,height=1.35cm]{fig/full_alphabets/7.pdf}
%     \end{center}
%   \caption{\textbf{Selected examples of generated fonts}. Examples generated by sampling a random latent representation $z$ and running the SVG decoder by conditioning on $z$ and all class labels. Each font character is selected as the best of 10 samples. Please see Figures \ref{fig:more-random-fonts} and \ref{fig:bad-random-fonts} in Appendix for additional examples.}
%     \label{fig:random-fonts}
% \end{figure*}

%\todo{not sure how much detail to go into here, and what should go in the Data and Model sections.}

We compiled a font dataset consisting of 14\,M examples. Individual font characters were normalized and converted into SVG format for training and evaluation. We trained a VAE and SVG decoder over $3$ epochs of the data and evaluated the results on a hold-out test split. Figures \ref{fig:svg} and \ref{fig:random-fonts} show selected results from the trained model, but please see Appendix (Figures \ref{fig:more-random-fonts} and \ref{fig:bad-random-fonts}) for more exhaustive samples highlighting successes and failures. What follows is an analysis of the representational ability of the model to learn and generate SVG specified fonts. 
%\ref{fig:smooth-latent}, \ref{fig:style-propagation} and \ref{fig:style-analogies} show selected results from the trained model but please see Appendix for more exhaustive samples highlighting successes (Figures \ref{fig:random-fonts}, \ref{fig:more-random-fonts}) and failures (Figure \ref{fig:bad-random-fonts}).

%We collect a dataset of fonts consisting of 14\,M examples that were scraped from the internet. We convert them from their original SFD format into SVG (which is a bijective conversion). We filter out glyphs that contain $>50$ commands (we found that the tail of the sequence-length distribution was long and cutting it short was key to achieve our results).

%These glyphs are first rendered and rasterized into a pixel representation, and used to train the image encoder and decoder models, which are also conditioned on the glyph's class. Then the SVG paths are converted into a sequence of vectors, each with 10 dimensions. The first 4 are a one-hot encoding of the command type (\svg{lineTo}, \svg{moveTo}, \svg{EOS}, etc) and the last 6 are the arguments to the given command. The SVG decoder is trained to output these commands autoregressively, while conditioned on the $z$ representation outputted by the pre-trained image encoder, as well as the glyph's class.

%-------------------------------------------------------------------------
\subsection{Learning a smooth, latent representation of font style}
\label{sec:smooth}
%\todo{should we add something about why do we care about perceptually smooth space? because we want it to be manipulable? or just because interesting (since we're introducing manipulations later)?}

\begin{figure*}[t!]
    \centering
    \begin{minipage}[t]{0.3\textwidth}
        \figsubtitle{Learned Latent Space}{\textwidth}%
        \vspace{0.0008\paperwidth}\\
        \includegraphics[trim={0 0 3in 0},clip=true,draft=false,width=\textwidth]{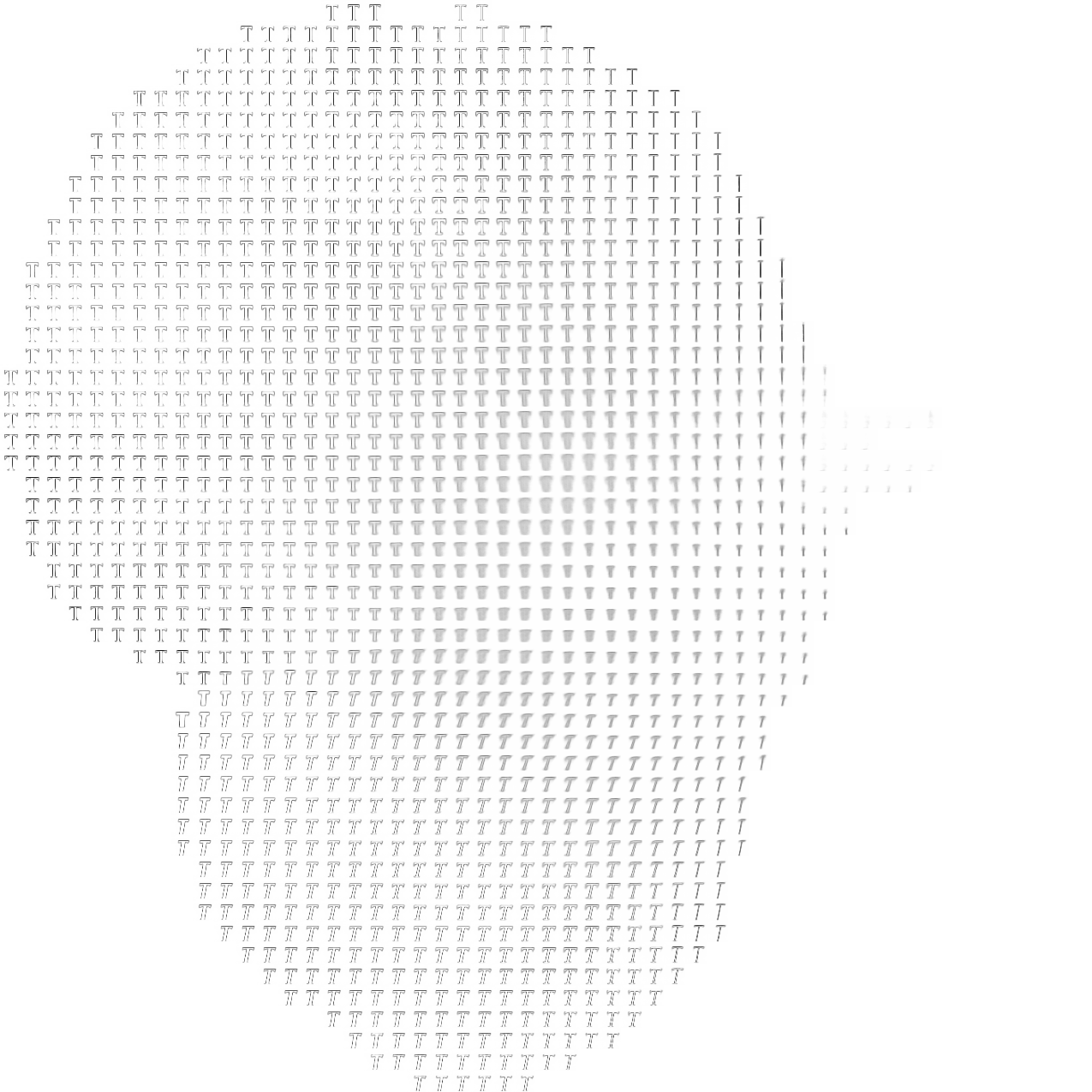}%
        \llap{\includegraphics[trim={0 0 1.618823in 0},clip=true,draft=false,width=\textwidth]{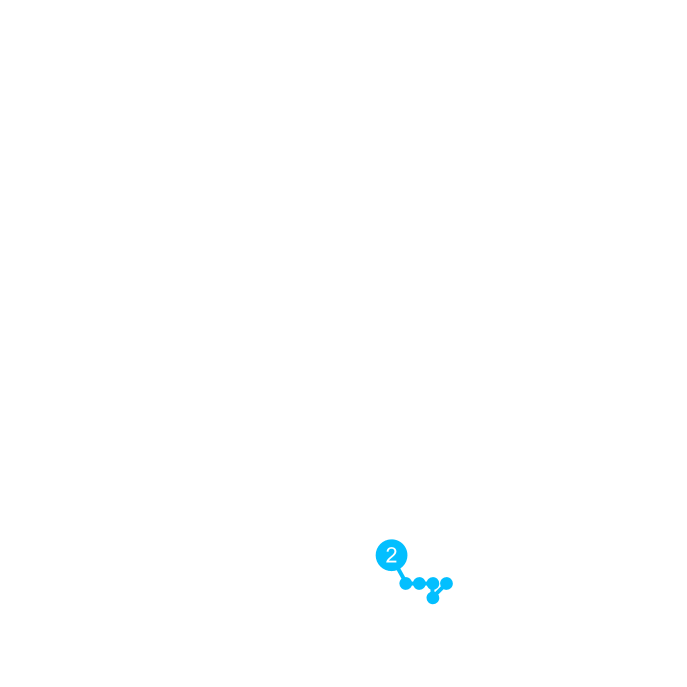}}%
        \llap{\includegraphics[trim={0 0 1.618823in 0},clip=true,draft=false,width=\textwidth]{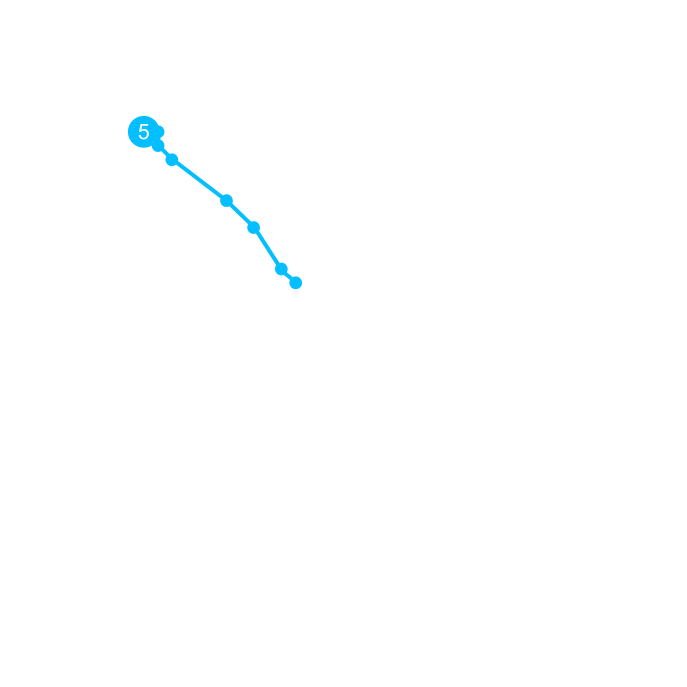}}%
        \llap{\includegraphics[trim={0 0 1.618823in 0},clip=true,draft=false,width=\textwidth]{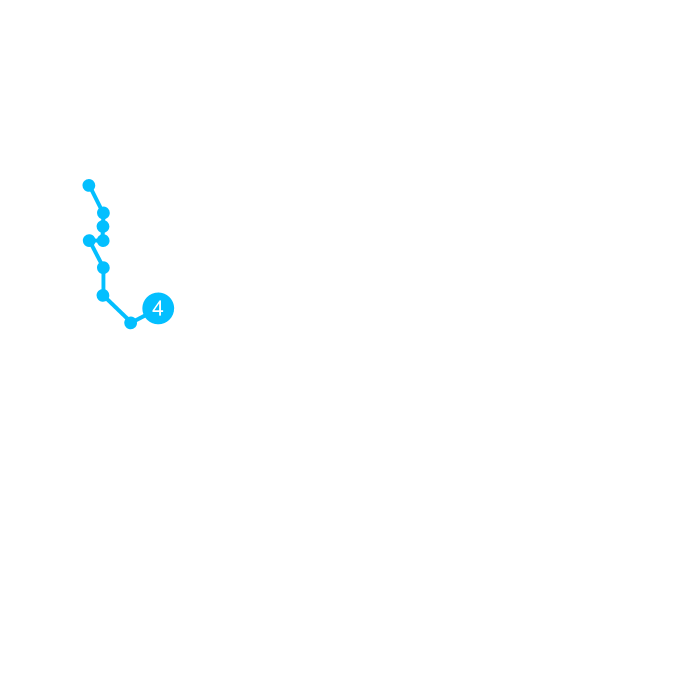}}%
        \llap{\includegraphics[trim={0 0 1.618823in 0},clip=true,draft=false,width=\textwidth]{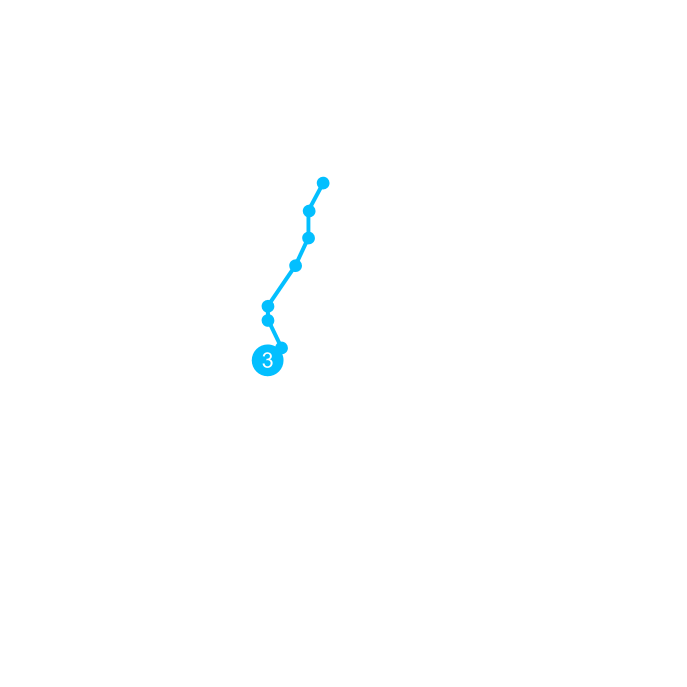}}%
        \llap{\includegraphics[trim={0 0 1.618823in 0},clip=true,draft=false,width=\textwidth]{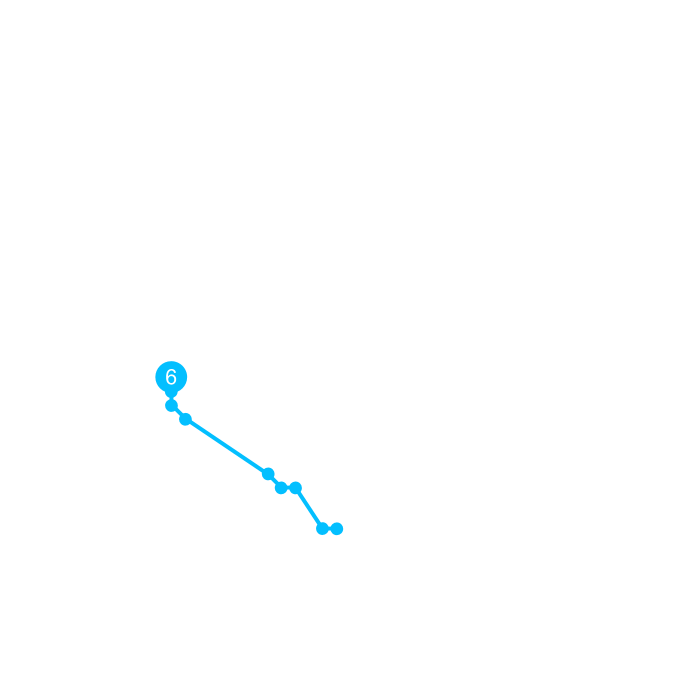}}%
        \llap{\includegraphics[trim={0 0 1.618823in 0},clip=true,draft=false,width=\textwidth]{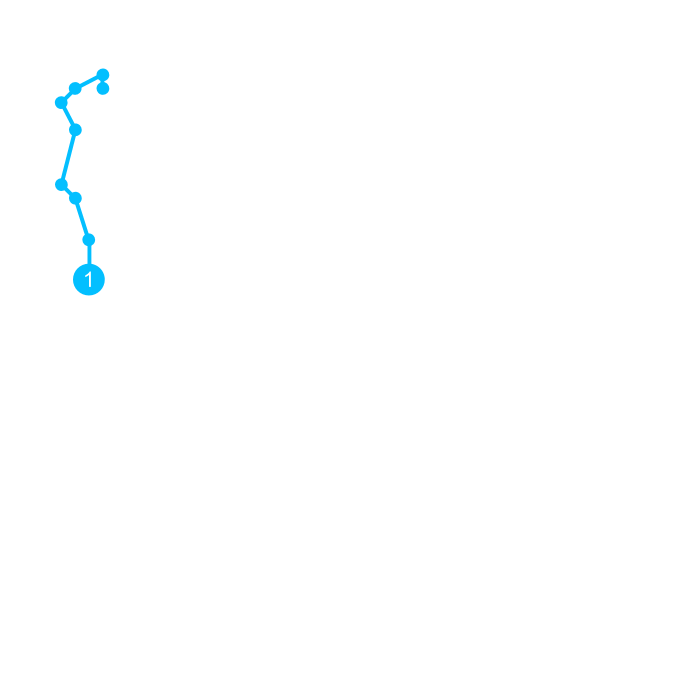}}%
        \llap{\includegraphics[trim={0 0 1.618823in 0},clip=true,draft=false,width=\textwidth]{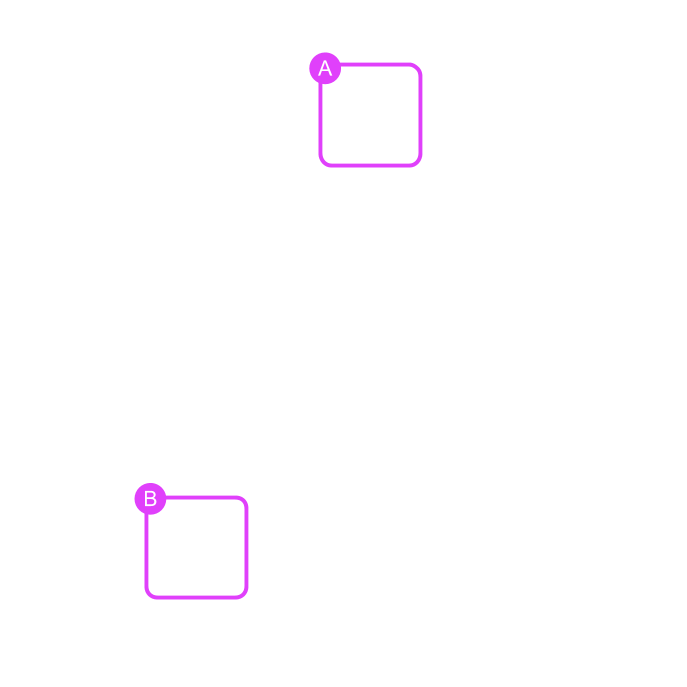}}%
    \end{minipage}
    \hspace{0.03\textwidth}%
    \begin{minipage}[t]{0.2\textwidth}
        \centering%
        \figsubtitle{Detail View}{\textwidth}%
        \vspace{0.008\paperwidth}\\
        \includegraphics[draft=false,width=0.8\textwidth]{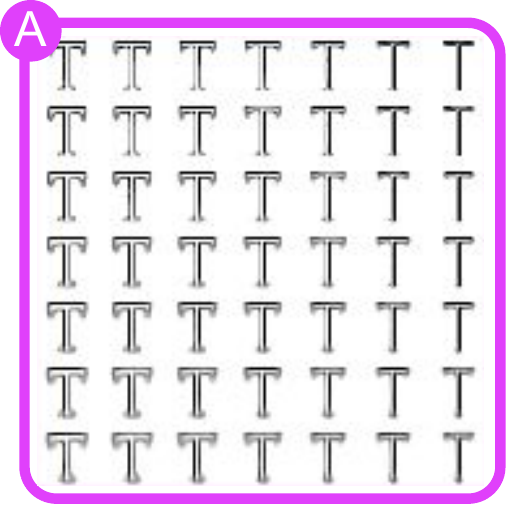}%
        \vspace{0.008\paperwidth}\\
        \includegraphics[draft=false,width=0.8\textwidth]{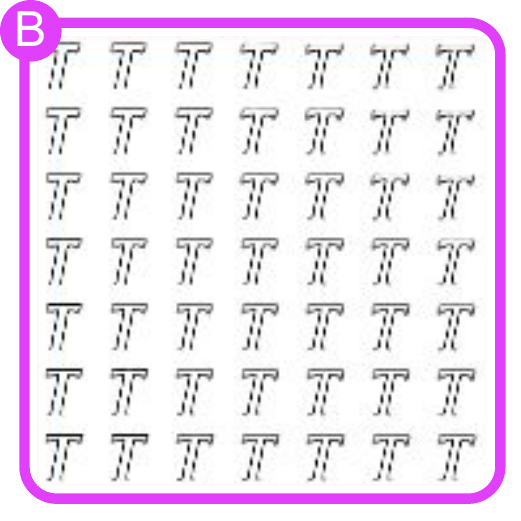}
    \end{minipage}%
    \hspace{0.03\textwidth}
    \begin{minipage}[t]{0.35\textwidth}
        \centering%
        \figsubtitle{Linear Interpolation Between Two Characters}{\textwidth}\\
        \vspace{0.01\textwidth}%
        \includegraphics[trim={0 0 0 0.19\textwidth},clip=true,draft=false,width=\textwidth]{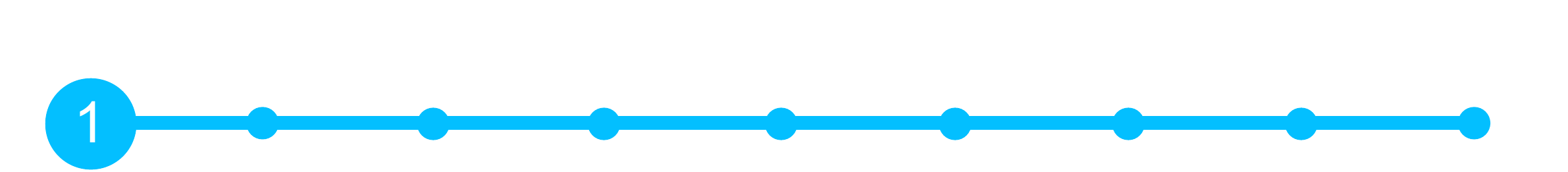}
        \includegraphics[trim={0 0 0 0.023\textwidth},clip=true,draft=false,width=\textwidth]{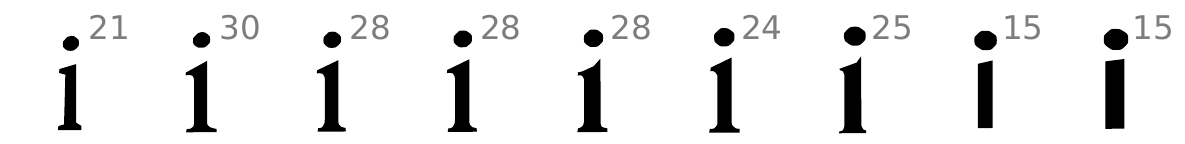}
        \includegraphics[trim={0 0 0 0.19\textwidth},clip=true,draft=false,width=\textwidth]{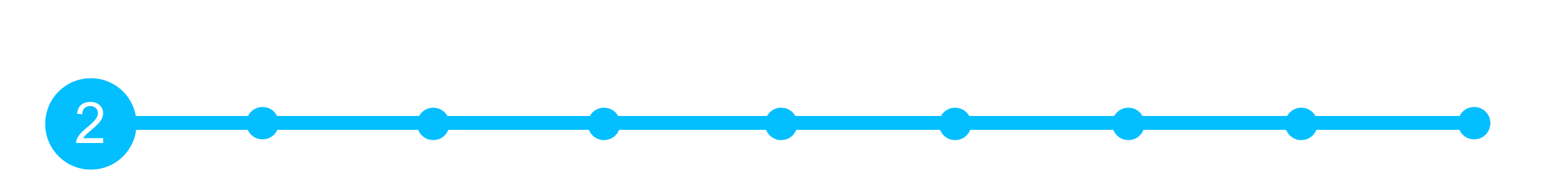}
        \includegraphics[trim={0 0 0 0.023\textwidth},clip=true,draft=false,width=\textwidth]{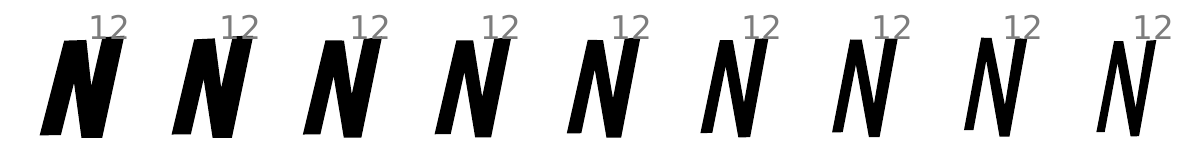}
        \includegraphics[trim={0 0 0 0.19\textwidth},clip=true,draft=false,width=\textwidth]{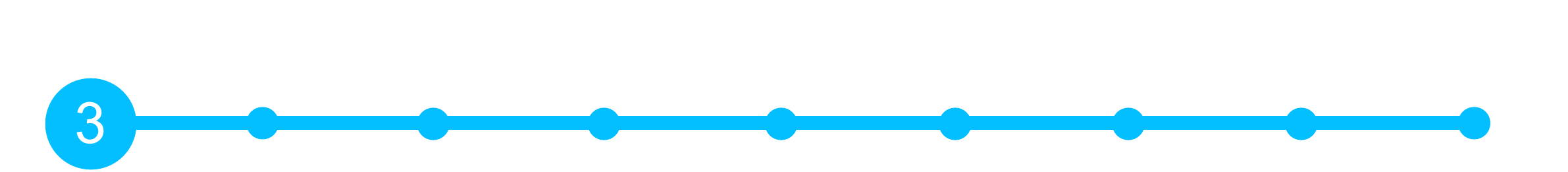}
        \includegraphics[trim={0 0 0 0.023\textwidth},clip=true,draft=false,width=\textwidth]{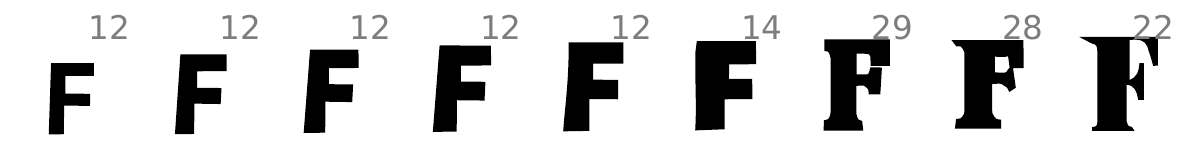}
        \includegraphics[trim={0 0 0 0.19\textwidth},clip=true,draft=false,width=\textwidth]{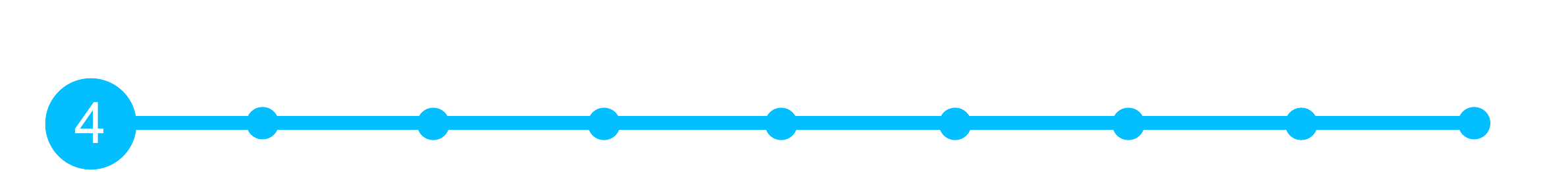}
        \includegraphics[trim={0 0 0 0.023\textwidth},clip=true,draft=false,width=\textwidth]{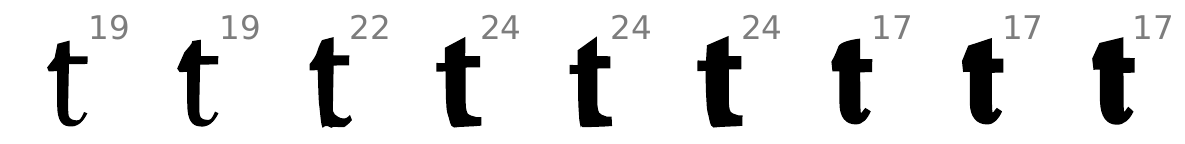}
        \includegraphics[trim={0 0 0 0.19\textwidth},clip=true,draft=false,width=\textwidth]{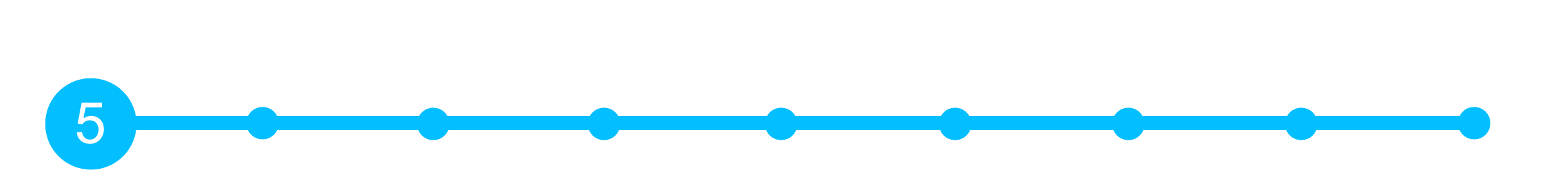}
        \includegraphics[trim={0 0 0 0.023\textwidth},clip=true,draft=false,width=\textwidth]{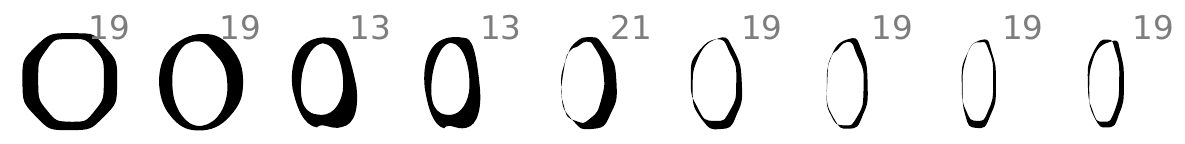}
        \includegraphics[trim={0 0 0 0.19\textwidth},clip=true,draft=false,width=\textwidth]{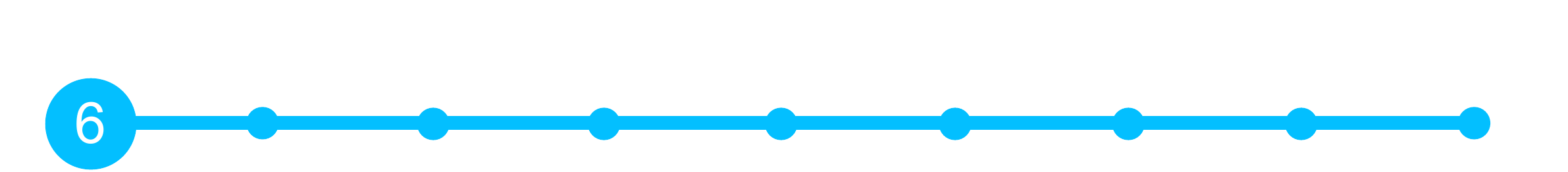}
        \includegraphics[trim={0 0 0 0.023\textwidth},clip=true,draft=false,width=\textwidth]{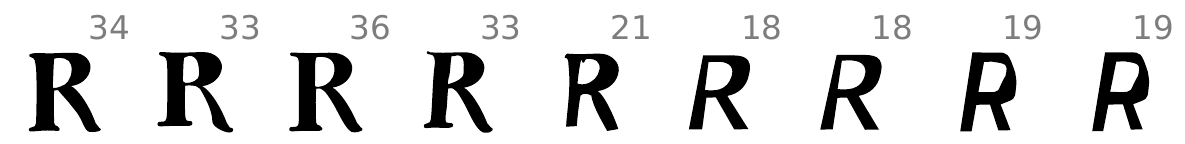}
    \end{minipage}
    \caption{\textbf{Learning a smooth, latent representation of font style}. UMAP visualization \cite{mcinnes2018umap} of the learned latent space $z$ across 1\,M examples (left). Purple boxes (A, B) provide a detail view of select regions. Blue lines (1-9) indicate \textit{linear} interpolations in the full latent space $z$ between two characters of the dataset. Points along these linear interpolations are rendered as SVG images. Number in upper-right corner indicates number of strokes in SVG rendering. Best viewed in digital color.}
    \label{fig:smooth-latent}
\end{figure*}

We first ask whether the proposed model may learn a latent representation for font style that is perceptually smooth and interpretable.  To address this question, we visualize the 32-dimensional font-style $z$ for 1\,M examples from the training set and reduce the dimensionality to 2 using UMAP \cite{mcinnes2018umap}. We discretize this 2D space and visualize the pixel-based decoding of the mean $z$ within each grid location (Figure \ref{fig:smooth-latent}). The purple boxes shows two separate locations of this manifold, where we note the smooth transitions of the characters: (A) represents a non-italics region, while (B) represents an italics one. Further, local directions within these regions also reveal visual semantics: within (A), from left-to-right we note a change in the amount of serif, while top-to-bottom highlights a change in boldness.

%In order to understand whether the learned latent corresponds to a perceptually smooth space, we visualize the learned latent VAE manifold by computing the 32-dimensional $z$ for 1\,M examples from the training set and reducing their dimensionality to 2 using UMAP \todo{(cite)}. We discretize this 2D space and, for each grid location, visualize the pixel decoding of the averaged 32-dim $z$s that lie in it (Figure \ref{fig:smooth-latent}, middle). Figure \ref{fig:smooth-latent} (left) shows two separate locations of this manifold, where we note their smooth transitions: purple represents a non-italics region, while red represents an italics one. Further, local directions within these regions also reveal visual semantics: within purple, from left-to-right we note a change in the amount of serif, while top-to-bottom highlights a change in boldness.

%\todo{transition with something like "but a latent that smoothly decodes pixel images would be nothing if it can't be used to decode SVGs that can also be visually smoothly interpolated. in order to see whether..." or something like "the above shows that the learned autoencoder can map visual features smoothly into the latent space. now we want to see if the sketch-rnn-like decoder can use this to smoothly decode SVGs"}

Next, we examine whether this smooth space translates into perceptually meaningful decodings of SVGs. We visualize linear interpolations between $z$ for pairs of SVGs from the dataset (Figure \ref{fig:smooth-latent}, 1-6). Note the smooth transition between the decoded SVGs, despite the fact that each SVG decoding is composed of a multitude of different commands. For instance, note that in the top row, each SVG is composed of 15-30 commands even though the perceptual representation appears quite smooth.

% \todo{Add number of commands for each item in the row of Figure 3}.
%(e.g.: the removal of serif in the second row's \classvalue{i}).

%\todo{conclude with "yeah our latent is smooth and can decode SVGs in that smoothly transition" perhaps some discussion about "this is possible because of our pretraining the latent. we tried learning autoencoding purely from command to command and with visual decoder as regularizer but we could not get it to work"}

%-------------------------------------------------------------------------
\subsection{Exploiting the latent representation for style propagation}
\label{sec:propagation}

\begin{figure}[t!]
    \begin{center}
        \includegraphics[draft=false,width=0.45\textwidth]{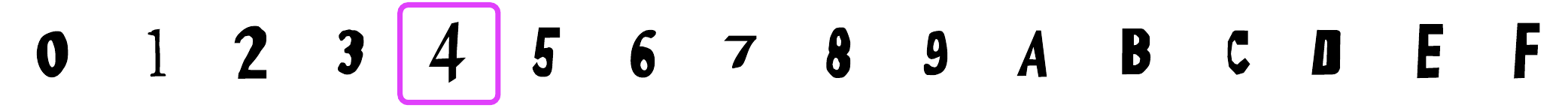}
        \includegraphics[draft=false,width=0.45\textwidth]{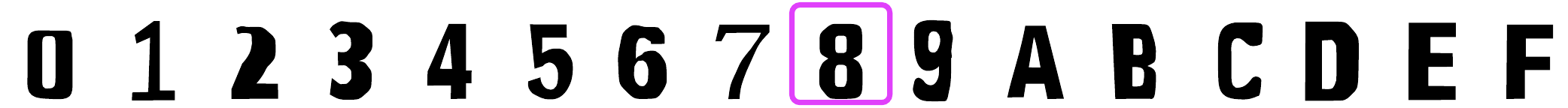}
        \includegraphics[draft=false,width=0.45\textwidth]{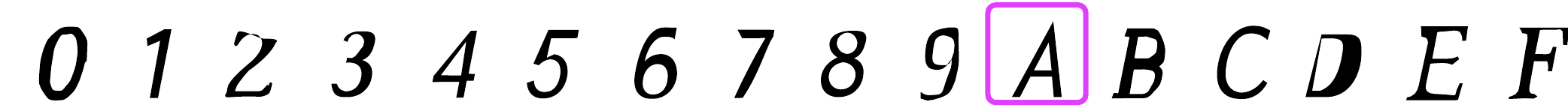}
        \includegraphics[draft=false,width=0.45\textwidth]{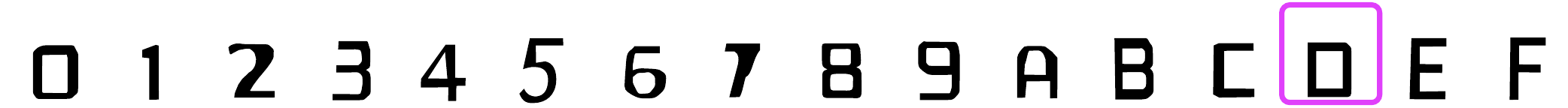}
        \includegraphics[draft=false,width=0.45\textwidth]{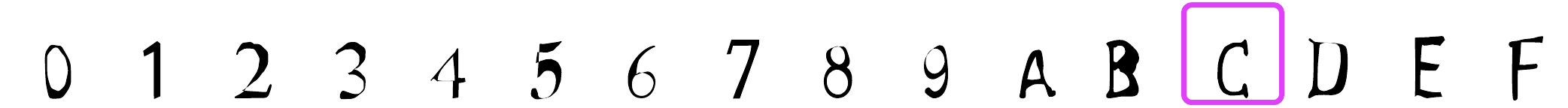}
        \includegraphics[draft=false,width=0.45\textwidth]{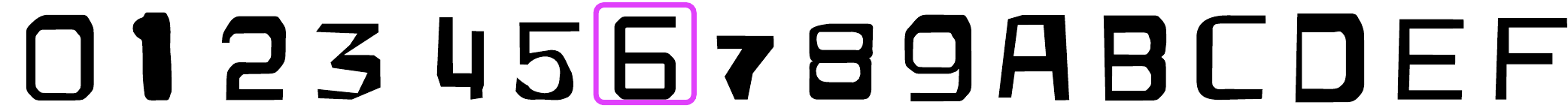}
        \includegraphics[draft=false,width=0.45\textwidth]{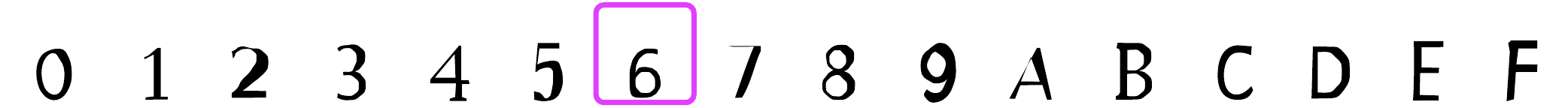}
        \includegraphics[draft=false,width=0.45\textwidth]{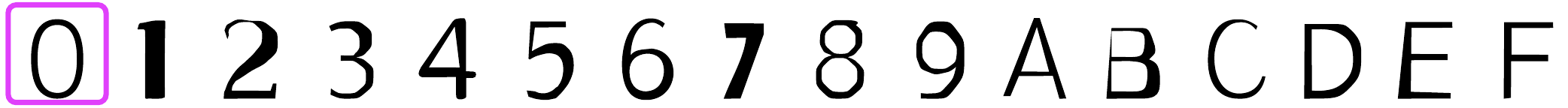}
        \includegraphics[draft=false,width=0.45\textwidth]{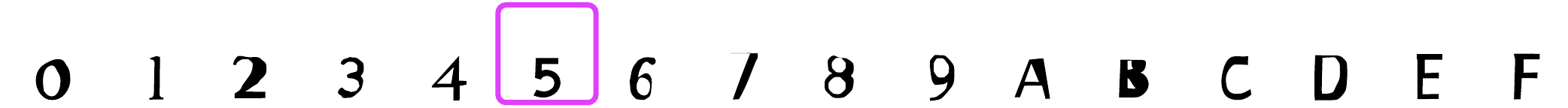}
        \includegraphics[draft=false,width=0.45\textwidth]{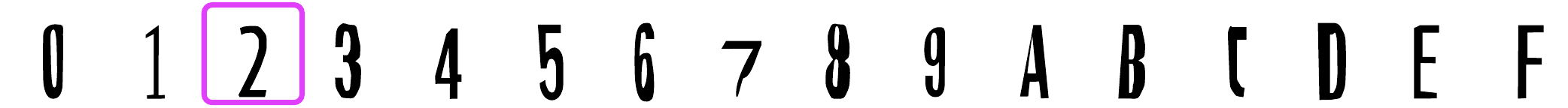}
        \includegraphics[draft=false,width=0.45\textwidth]{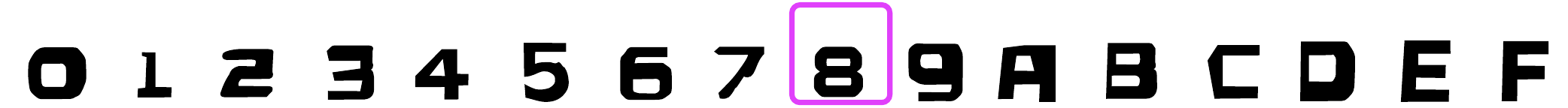}
        \includegraphics[draft=false,width=0.45\textwidth]{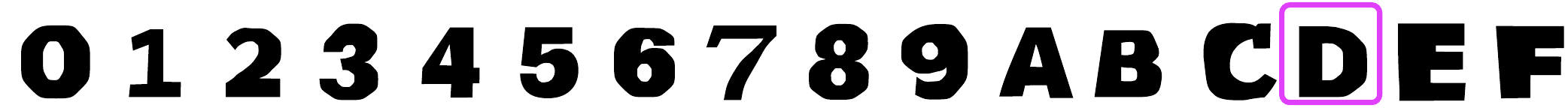}
        \includegraphics[draft=false,width=0.45\textwidth]{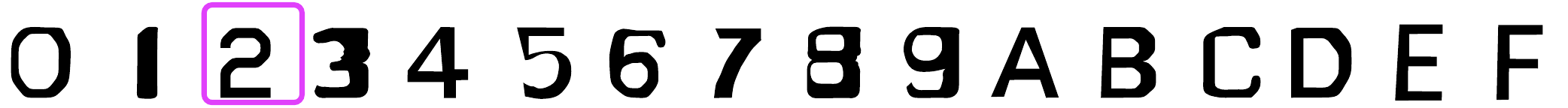}
        \includegraphics[draft=false,width=0.45\textwidth]{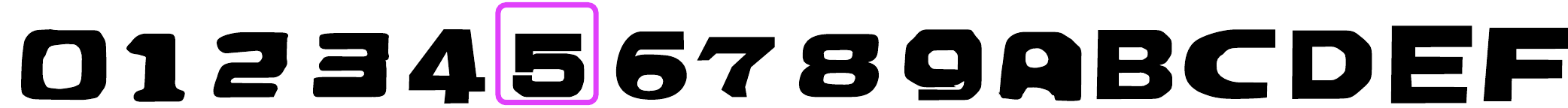}

    \end{center}
    \caption{\textbf{Exploiting the latent representation for style propagation}. A single character may provide sufficient information for reconstructing the rest of a font set. The latent representation $z$ for a font is computed from a {\it single} character (purple box) and SVG images are generated for other characters from $z$.} %\caption{\textbf{Exploiting the latent representation for style propagation}. Because the latent representation if class-agnostic, we can transfer the style of a glyph to other classes by: 1) acquiring the style of a glyph (highlighted in blue) via the image encoder 2) decoding different classes by giving different class labels to the SVG decoder (other glyphs in the same row). Style correspondences emerge in an unsupervised way.}
    \label{fig:style-propagation}
\end{figure}

\begin{figure}[t!]
    \begin{center}
        \includegraphics[draft=false,width=0.45\textwidth]{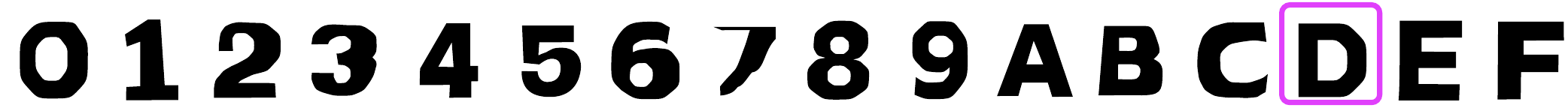}
        \includegraphics[draft=false,width=0.45\textwidth]{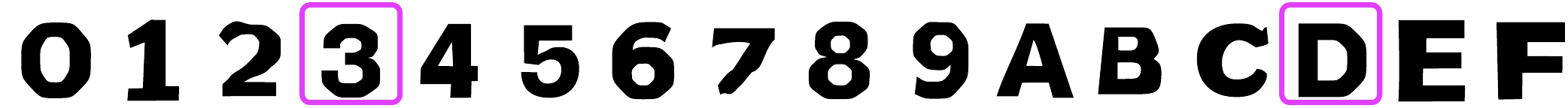}
        \includegraphics[draft=false,width=0.45\textwidth]{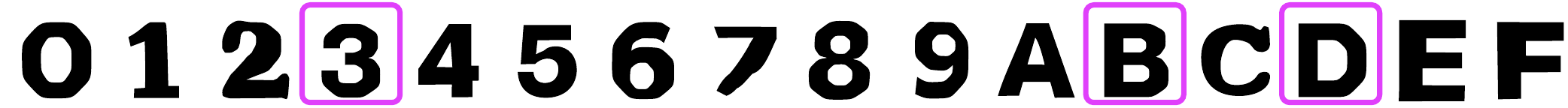}
        \includegraphics[draft=false,width=0.45\textwidth]{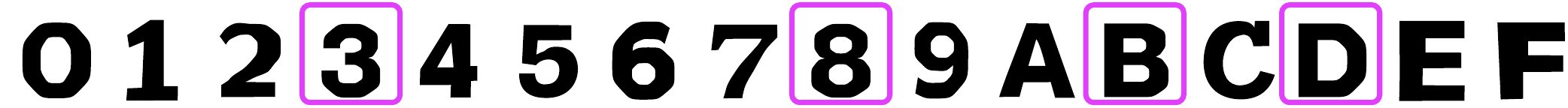}
        \includegraphics[draft=false,width=0.45\textwidth]{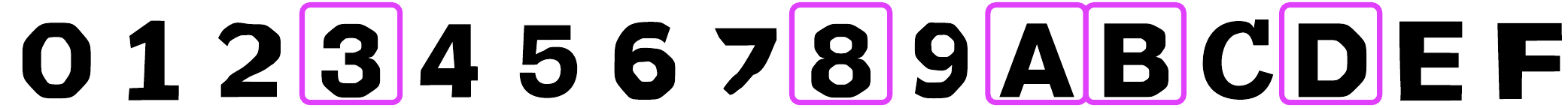}%
        \vspace{0.01\textwidth}
        \includegraphics[draft=false,width=0.45\textwidth]{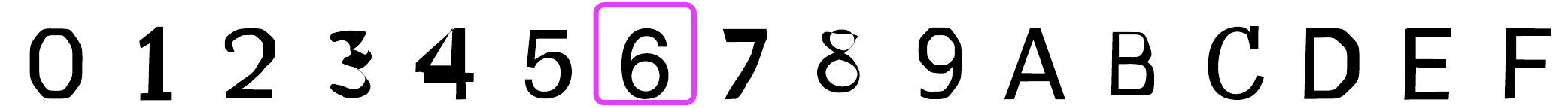}
        \includegraphics[draft=false,width=0.45\textwidth]{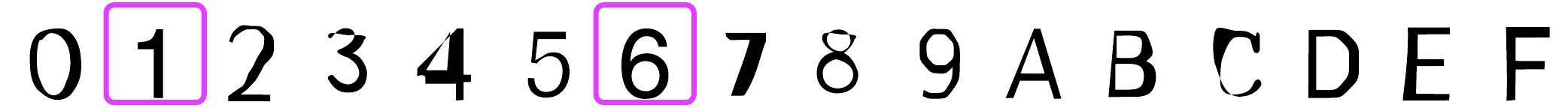}
        \includegraphics[draft=false,width=0.45\textwidth]{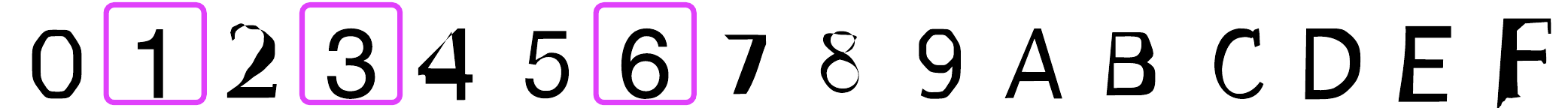}
        \includegraphics[draft=false,width=0.45\textwidth]{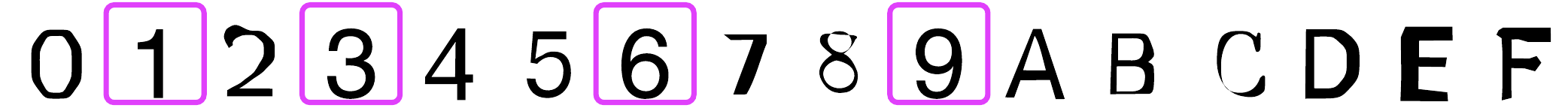}
        \includegraphics[draft=false,width=0.45\textwidth]{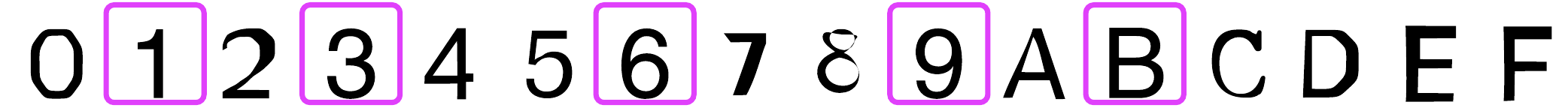}%
        \vspace{0.01\textwidth}
        % \includegraphics[draft=false,width=0.45\textwidth]{fig/multi_cond/20_middle.pdf}
        % \includegraphics[draft=false,width=0.45\textwidth]{fig/multi_cond/21_middle.pdf}
        % \includegraphics[draft=false,width=0.45\textwidth]{fig/multi_cond/22_middle.pdf}
        % \includegraphics[draft=false,width=0.45\textwidth]{fig/multi_cond/23_middle.pdf}
        % \includegraphics[draft=false,width=0.45\textwidth]{fig/multi_cond/24_middle.pdf}
        % \includegraphics[draft=false,width=0.45\textwidth]{fig/multi_cond/25_middle.pdf}
        % \includegraphics[draft=false,width=0.45\textwidth]{fig/multi_cond/26_middle.pdf}
        % \includegraphics[draft=false,width=0.45\textwidth]{fig/multi_cond/27_middle.pdf}
        % \includegraphics[draft=false,width=0.45\textwidth]{fig/multi_cond/28_middle.pdf}
        % \includegraphics[draft=false,width=0.45\textwidth]{fig/multi_cond/29_middle.pdf}
        % \includegraphics[draft=false,width=0.45\textwidth]{fig/multi_cond/30_middle.pdf}
        % \includegraphics[draft=false,width=0.45\textwidth]{fig/multi_cond/31_middle.pdf}
        % \includegraphics[draft=false,width=0.45\textwidth]{fig/multi_cond/32_middle.pdf}
        % \todo{pick more than 1 to highlight}
        \includegraphics[draft=false,width=0.45\textwidth]{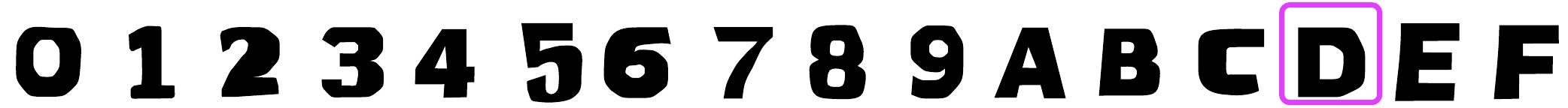}
        \includegraphics[draft=false,width=0.45\textwidth]{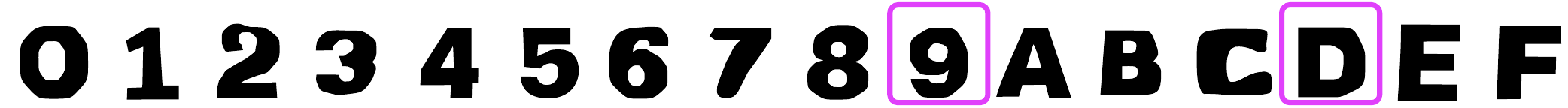}
        \includegraphics[draft=false,width=0.45\textwidth]{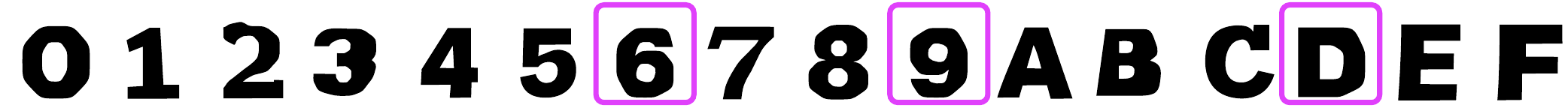}
        \includegraphics[draft=false,width=0.45\textwidth]{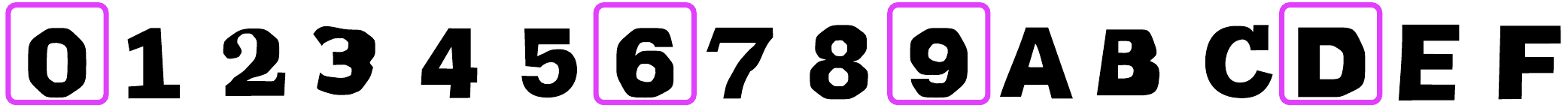}
        \includegraphics[draft=false,width=0.45\textwidth]{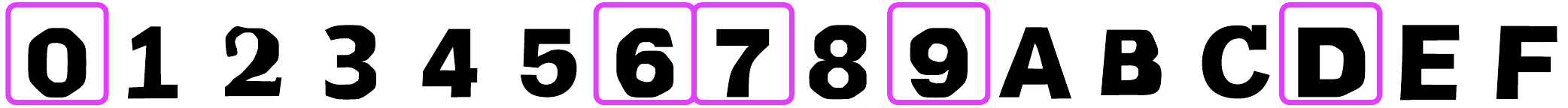}%
        \vspace{0.005\textwidth}
        \includegraphics[draft=false,width=0.23\textwidth]{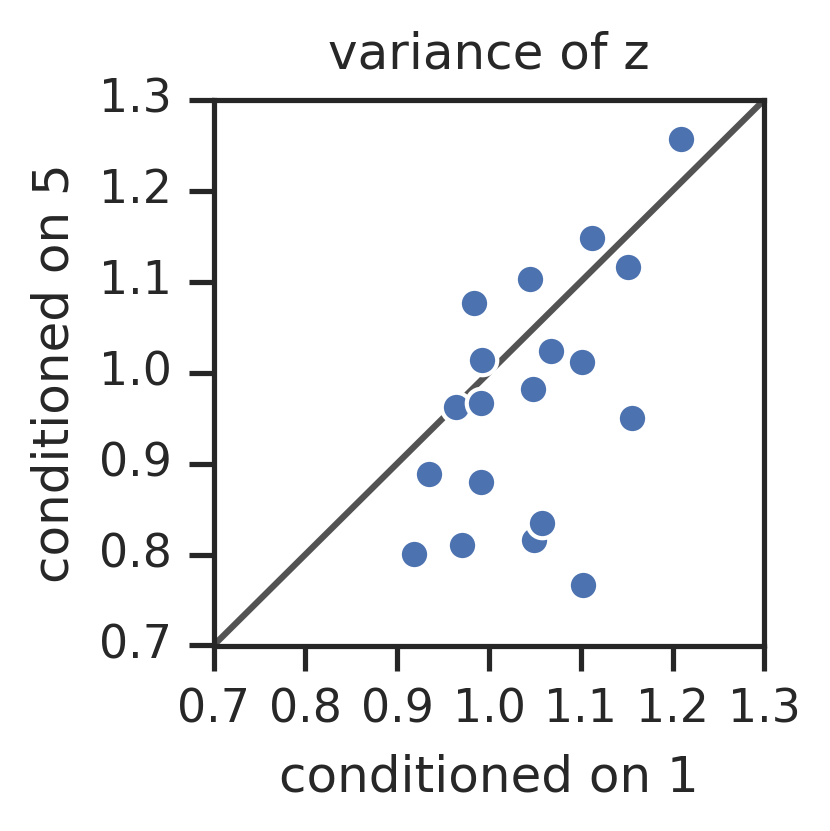}
        \vspace{-0.02\textwidth}
    \end{center}
    \caption{\textbf{Conditioning on increasing numbers of characters improves style propagation.} Top: Layout follows Figure \ref{fig:style-propagation}. The average latent representation $z$ for a font is computed from a set of characters (purple boxes) and SVG images are generated for other characters from $z$. Note that increasing number of characters (purple boxes) improves the consistency and quality of the style propagation. Bottom: For all generated characters, we calculate a corresponding $z$ and measure the variance of $z$ over all generated characters within a font. Lower variance in $z$ indicates a more visually consistent font style. Each dot corresponds to the observed variance in $z$ when conditioned on $1$ or $5$ characters. Note that most fonts contain higher consistency (i.e. lower variance) when conditioned on more characters.}
    %\caption{\todo{\textbf{Conditioning on multiple icons for higher confidence style representations}. Each row of characters is computed by averaging the $z$s of the ground-truth characters highlighted in red. By conditioning on multiple characters, the $z$ estimate of font style becomes more precise, which allows for higher quality and more style-consistent outputs. This could similarly be done iteratively by selecting desirable outputs at each step.}}
    \label{fig:multi-style-propagation}
\end{figure}

Because the VAE is conditioned on the class label, we expect that the latent representation $z$ would only encode the font style with minimal class information \cite{kingma2014semi}. We wish to exploit this model structure to perform style propagation across fonts. In particular, we ask whether a {\it single character} from a font set is sufficient to infer the rest of the font set in a visually plausible manner \cite{rehling2001letter, hofstadter1993letter}.

To perform this task, we calculate the latent representation $z$ for a single character and condition the SVG decoder on $z$ as well as the label for all other font characters (i.e. \classvalue{0}-\classvalue{9}, \classvalue{a}-\classvalue{z}, \classvalue{A}-\classvalue{Z}). Figure \ref{fig:style-propagation} shows the results of this experiment. For each row, $z$ is calculated from the character in the red box. The other characters in that row are generated from the SVG decoder conditioned on $z$.

%In Figure \ref{fig:style-propagation}, glyphs in same row are conditioned on the same $z$, obtained from the glyph highlighted in blue, but different classes. 
We observe a perceptually-similar style consistently within each row \cite{rehling2001letter, hofstadter1993letter}.  Note that there was no requirement during training that the same point in latent space would correspond to a perceptually similar character across labels -- that is, the consistency across class labels was learned in an unsupervised manner \cite{rehling2001letter, hofstadter1993letter}. Thus, a single value of $z$ seems to correspond to a perceptually-similar set of characters that resembles a plausible fontset.

Additionally, we observe a large amount of style variety across rows (i.e. different $z$) in Figure \ref{fig:style-propagation}. The variety indicates that the latent space $z$ is able to learn and capture a large diversity of styles observed in the training set as observed in Figure \ref{fig:smooth-latent}.

Finally, we also note that for a given column the decoded glyph does indeed belong to the class that was supplied to the SVG decoder. These results indicate that $z$ encodes style information consistently across different character labels, and that the proposed model largely disentangles class label from style. 

A natural extension to this experiment is to ask if we could systematically improve the quality of style propagation by employing more then a single character. We address this question by calculating the latent representation $z$ for multiple characters and employ the average $z$ for style propagation to a new set of characters (Figure \ref{fig:multi-style-propagation}). We observe a systematic improvement in both style consistency and quality of individual icon outputs as one conditions on increasing numbers of characters.

To quantify this improvement in style consistency, we render the generated characters and calculate the associated style $z$ for each character. If the method of style propagation were perfectly self-consistent, we would expect that the $z$ across all generated characters would be identical. If, however, the style propagation were not consistent, the inferred $z$ would vary across each of the generated characters. To calculate the observed improvement, we measure the variance of $z$ across all generated characters for each of 19 fonts explored with this technique when conditioned on on $1$ or $5$ characters (Figure \ref{fig:multi-style-propagation}, bottom). Indeed, we observe that conditioning on more characters generally decreases the variance of the generated styles, indicating that this procedure improves style consistency.
Taken together, we suspect that these results on style progagation suggest a potential direction for providing iterative feedback to humans for synthesizing new fonts (see Discussion).

\subsection{Building style analogies with the learned representation}
\label{sec:analogies}

\begin{figure}[t!]
    \begin{center}
    \begin{minipage}[l]{0.15\textwidth}
        \includegraphics[draft=false,width=0.9\textwidth]{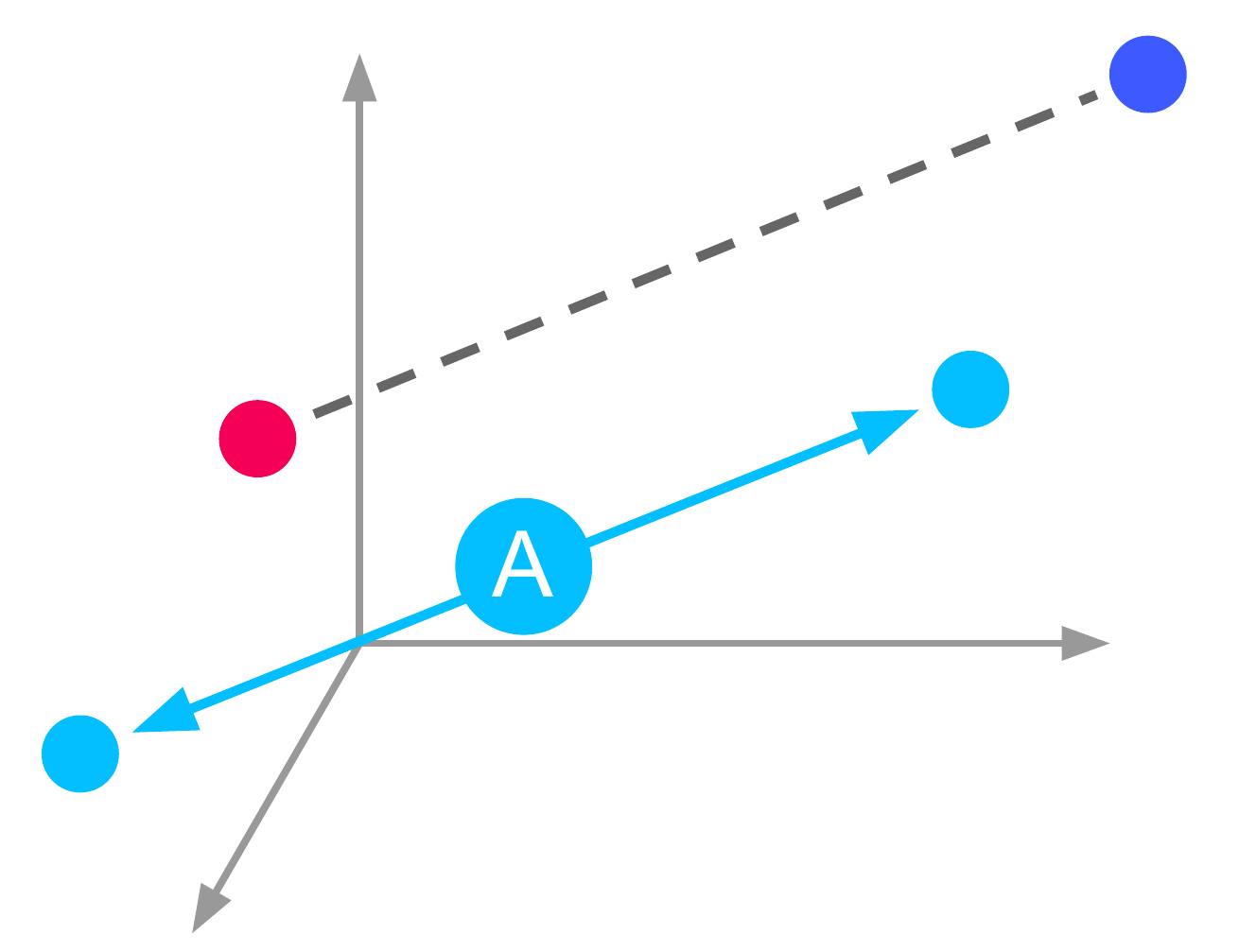}%
    \end{minipage}
    \begin{minipage}[l]{0.15\textwidth}
        \includegraphics[draft=false,width=\textwidth]{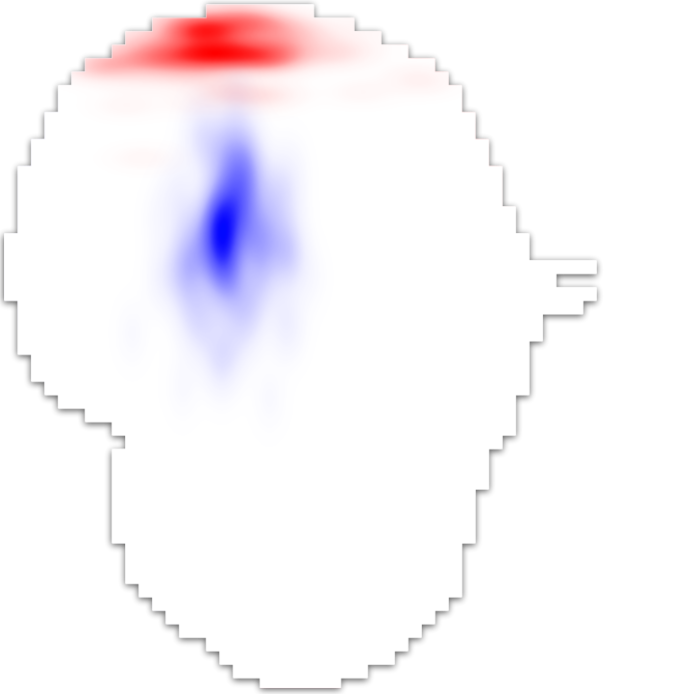}%
    \end{minipage}%
    \begin{minipage}[l]{0.3\textwidth}
        \begin{center}
            % \text{bold$\rightarrow$}\vspace{0.05in}
            \includegraphics[draft=false,width=\textwidth]{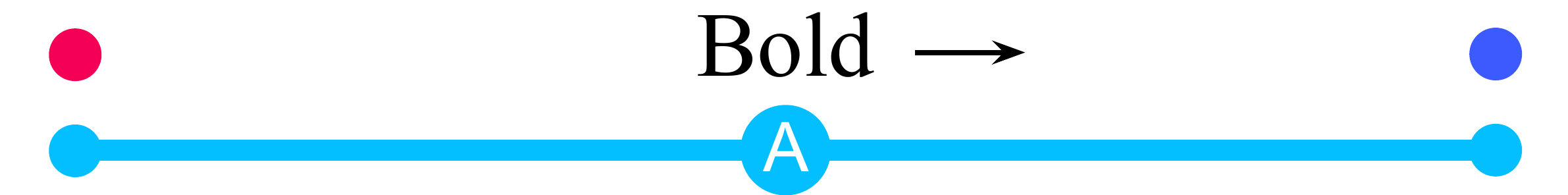}
            \includegraphics[draft=false,width=\textwidth]{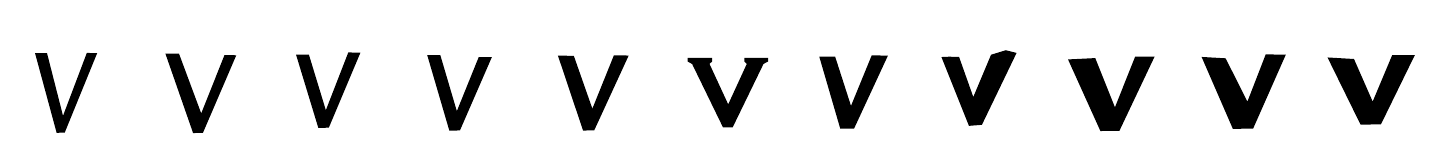}
            \includegraphics[draft=false,width=\textwidth]{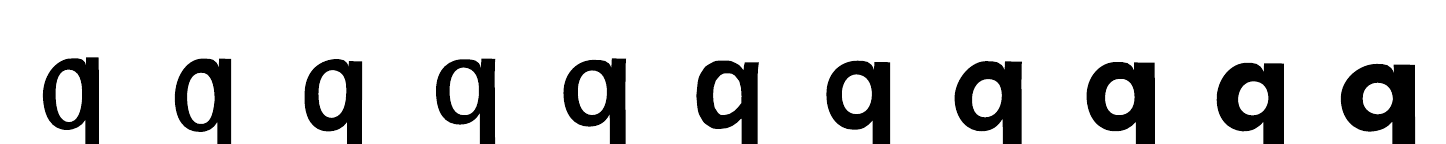}
            \includegraphics[draft=false,width=\textwidth]{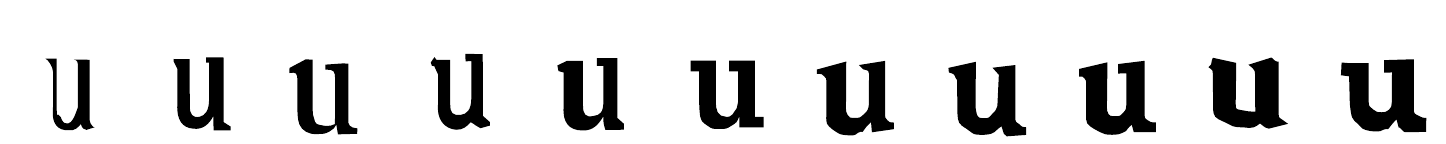}
            \includegraphics[draft=false,width=\textwidth]{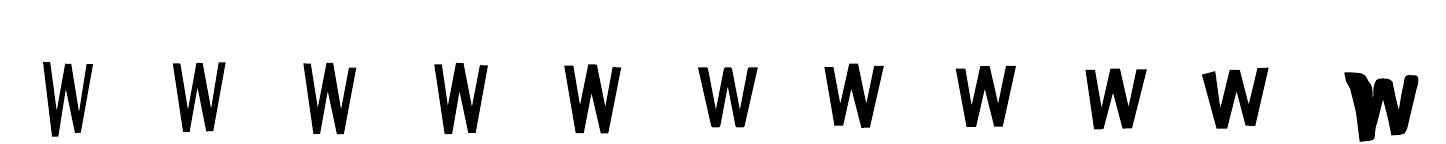}
        \end{center}
    \end{minipage}%
    \vspace{0.2in}
    \begin{minipage}[l]{0.15\textwidth}
        \includegraphics[draft=false,width=\textwidth]{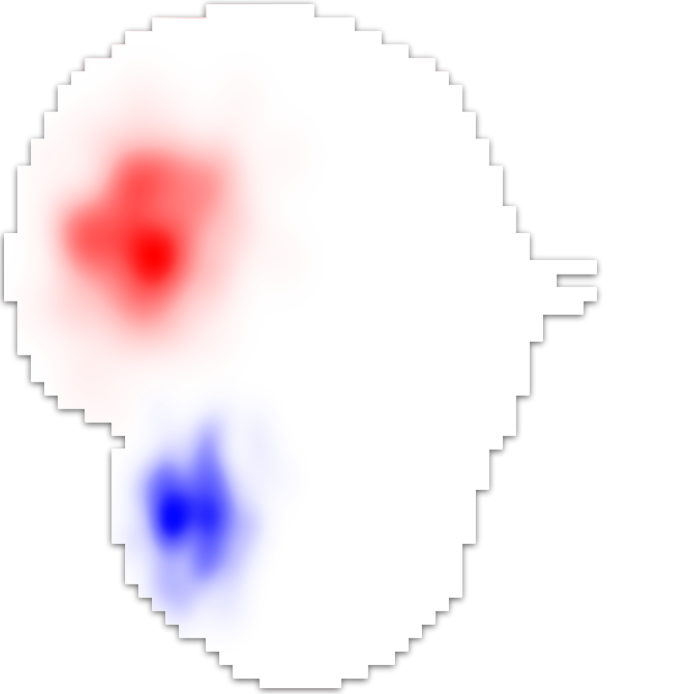}%
    \end{minipage}%
    \begin{minipage}[l]{0.3\textwidth}
        \begin{center}
            % \text{italic$\rightarrow$}\vspace{0.05in}
            \includegraphics[draft=false,width=\textwidth]{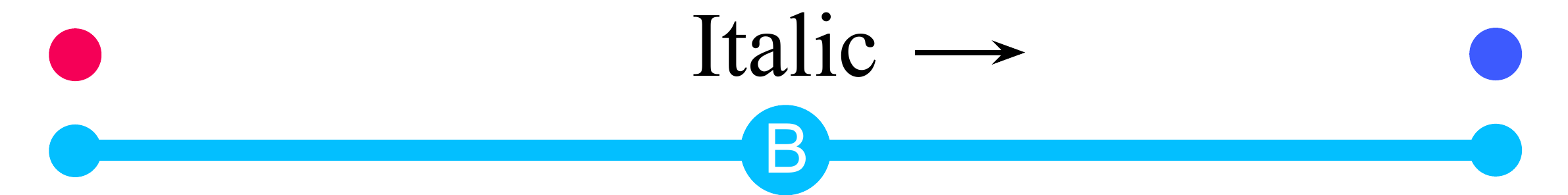}
            \includegraphics[draft=false,width=\textwidth]{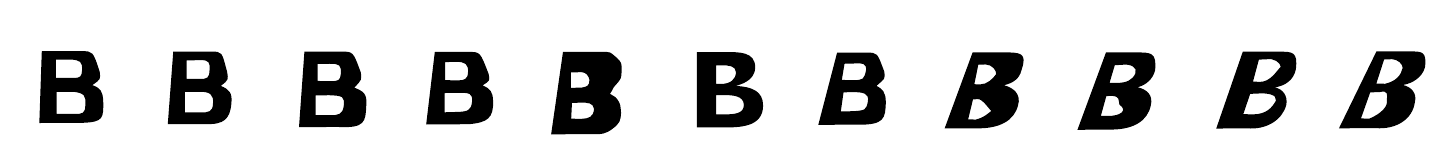}
            \includegraphics[draft=false,width=\textwidth]{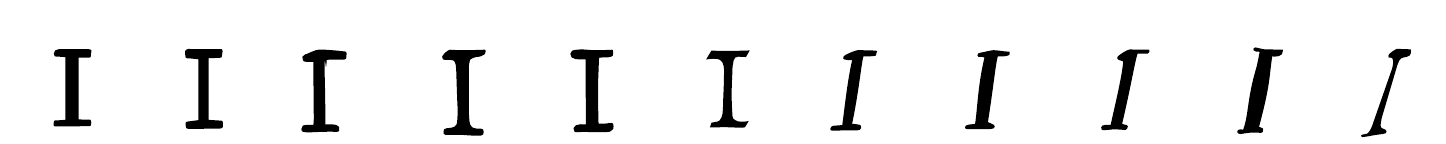}
            \includegraphics[draft=false,width=\textwidth]{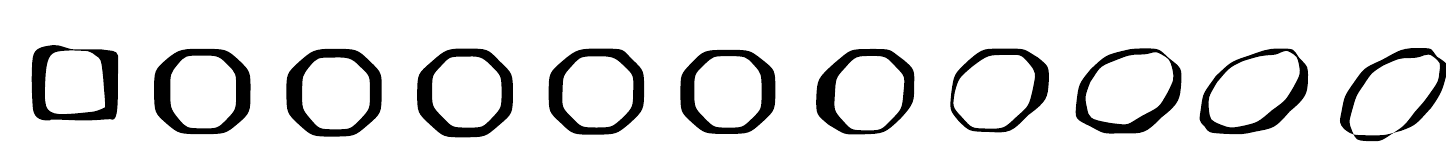}
            \includegraphics[draft=false,width=\textwidth]{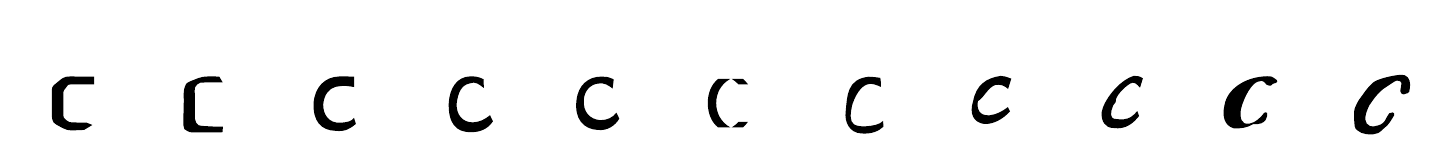}
        \end{center}
    \end{minipage}%
    \vspace{0.2in}
    \begin{minipage}[l]{0.15\textwidth}
        \includegraphics[draft=false,width=\textwidth]{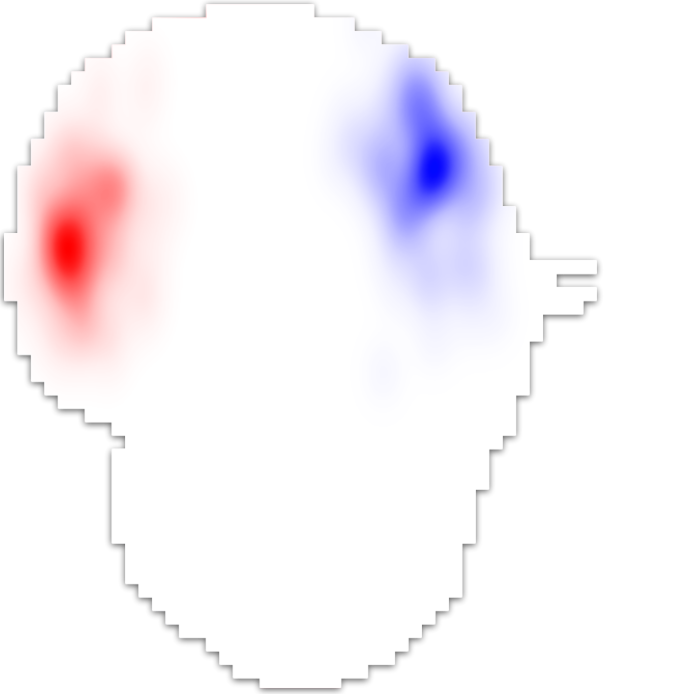}%
    \end{minipage}%
    \begin{minipage}[l]{0.3\textwidth}
        \begin{center}
            % \text{condensed$\rightarrow$}\vspace{0.05in}
            \includegraphics[draft=false,width=\textwidth]{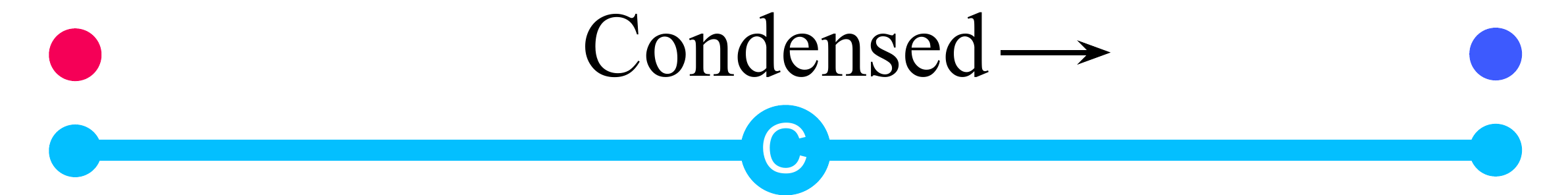}
            \includegraphics[draft=false,width=\textwidth]{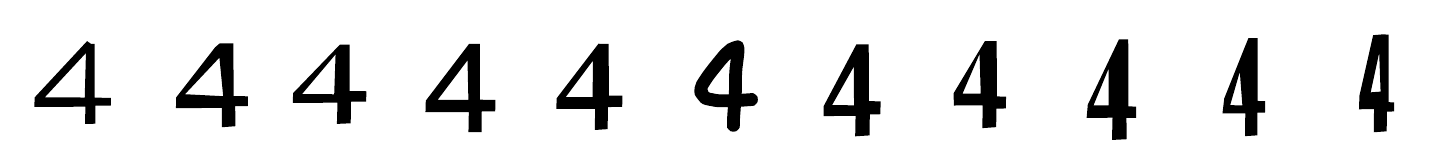}
            \includegraphics[draft=false,width=\textwidth]{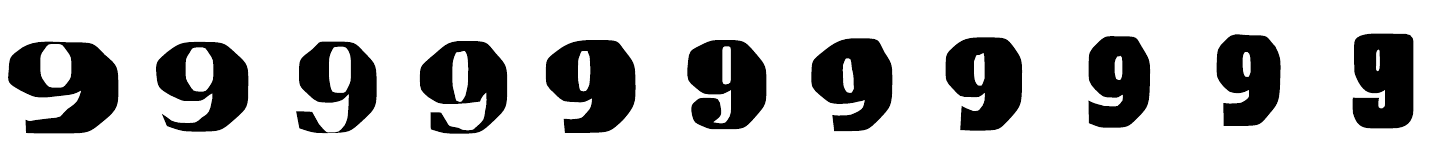}
            \includegraphics[draft=false,width=\textwidth]{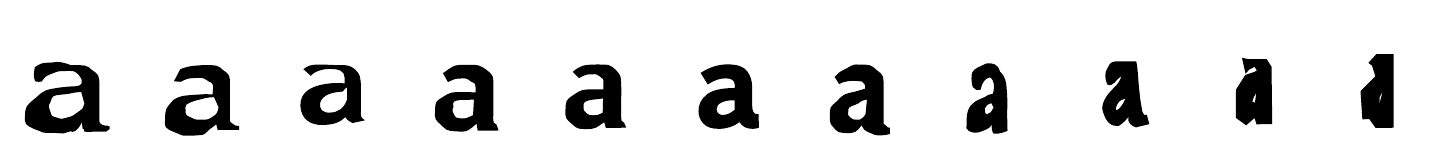}
            \includegraphics[draft=false,width=\textwidth]{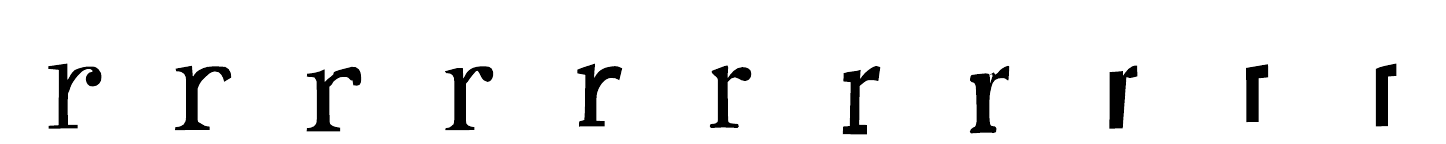}
        \end{center}
    \end{minipage}%
    \end{center}
    \caption{\textbf{Building style analogies with the learned representation}. Semantically meaningful directions may be identified for globally altering font attributes. Top row: Bold (blue) and non-bold (red) regions of latent space (left) provide a vector direction that may be added to arbitrary points in latent space (A) for decreasing or increasing the strength of the attribute. Middle and bottom rows: Same for italics (B) and condensed (C).}
    %\caption{\textbf{Building style analogies with the learned representation}. Semantically meaningful directions of the latent space are surfaced by selecting random data points that correspond to positive and negative examples of a concept (e.g.: bold and non-bold glyphs, visualized as red and blue heatmaps at the top) and finding the direction between their means. Then, any example (yellow) can be modified by adding (teal) or subtracting (maroon) the concept direction from its latent representation. Here we visualize the boldness direction (top), italics direction (center) and condensed direction (bottom).}
    \label{fig:style-analogies}
\end{figure}

Given that the latent style is perceptually smooth and aligned across class labels, we next ask if we may find semantically meaningful directions in this latent space. In particular, we ask whether these semantically meaningful directions may permit global manipulations of font style.
%With the knowledge that the style latent is both visually smooth, and aligned across classes, it's possible there are directions in the latent manifold that encode high level concept that we are interested in manipulating.

Inspired by the work on word vectors \cite{mikolov2013distributed}, we ask whether one may identify analogies for organizing the space of font styles (Figure \ref{fig:style-analogies}, top). To address this question, we select positive and negative examples for semantic concepts of organizing fonts (e.g. bold, italics, condensed) and identify regions in latent space corresponding to the presence or absence of this concept (Figure \ref{fig:style-analogies}, left, blue and red, respectively). We compute the average $z_{red}$ and $z_{blue}$, and define the concept direction $c = z_{blue} - z_{red}$.

We test if these directions are meaningful by taking an example font style $z^{*}$ from the dataset (Figure \ref{fig:style-analogies}, right, yellow), and adding (or subtracting) the concept vector $c$ scaled by some parameter $\alpha$.  Finally, we compute the SVG decodings for $z^{*} + \alpha c$ across a range of $\alpha$. 

Figure \ref{fig:style-analogies} (right) shows the resulting fonts. Note that across the three properties examined, we observe a smooth interpolation in the direction of the concept  modeled (e.g.: first row \classvalue{v} becomes increasingly bold from left to right). We take these results to indicate that one may interpret semantically meaningful directions in the latent space. Additionally, these results indicate that one may find directions in the latent space to globally manipulate font style.

%In order to see if these style analogies could be inferred from the latent $z$, we select positive (Figure \ref{fig:style-analogies}, left, blue) and negative (red) examples of certain concepts (bold, italics, condensed). We then compute the average $z_{red}$ and $z_{blue}$, and define the concept direction $c = z_{blue} - z_{red}$. Finally, we test if these directions are meaningful by taking a random example from the dataset (Figure \ref{fig:style-analogies}, right, yellow), and adding/subtracting the concept vector $c$ scaled by some scalar value. The SVG decodings for various scales can be seen in Figure \ref{fig:style-analogies} (right). They show smooth interpolation in the direction of the concept direction being modelled (e.g.: first row \classvalue{v} becomes increasingly bold from left to right). This indicates that our style analogies are meaningful and useful for manipulating font styles.

%-------------------------------------------------------------------------
\subsection{Quantifying the quality of the learned representations}
\label{sec:quantifying}

\begin{figure*}[t!]
    \centering
    \subfloat[\label{fig6:a}]{
    \begin{minipage}[t]{0.25\textwidth}
        \figsubtitle{Assessing Generalization}{\textwidth}
        \vspace{0.005in}\\
        \includegraphics[draft=false,width=\textwidth]{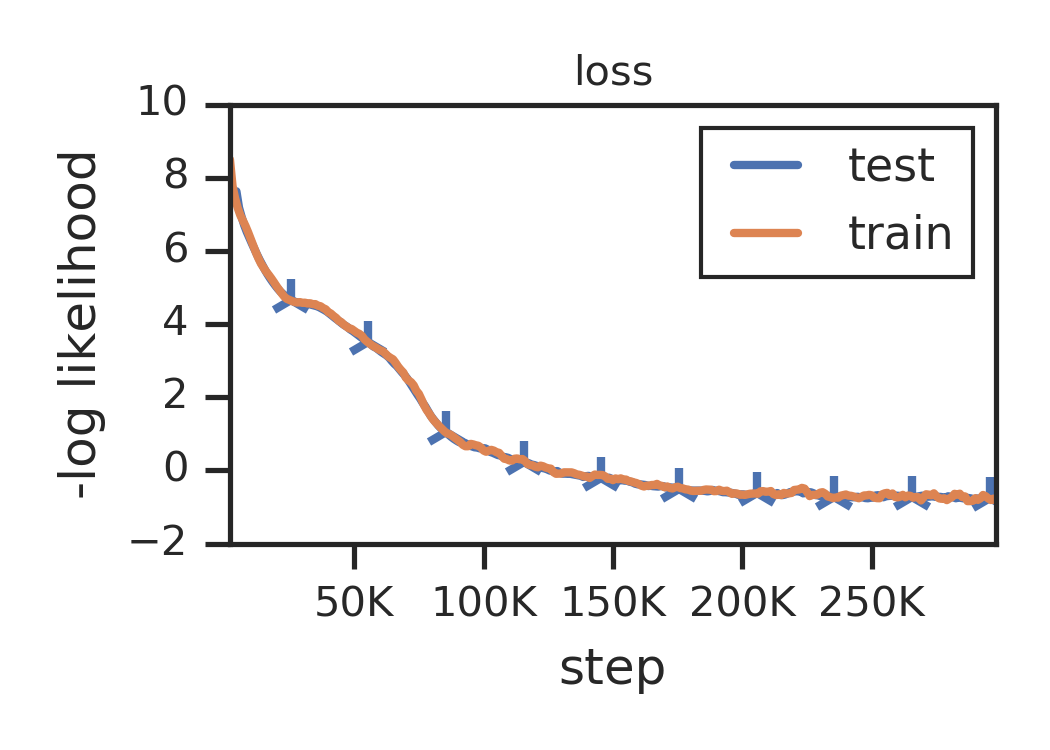}
        \includegraphics[draft=false,width=\textwidth]{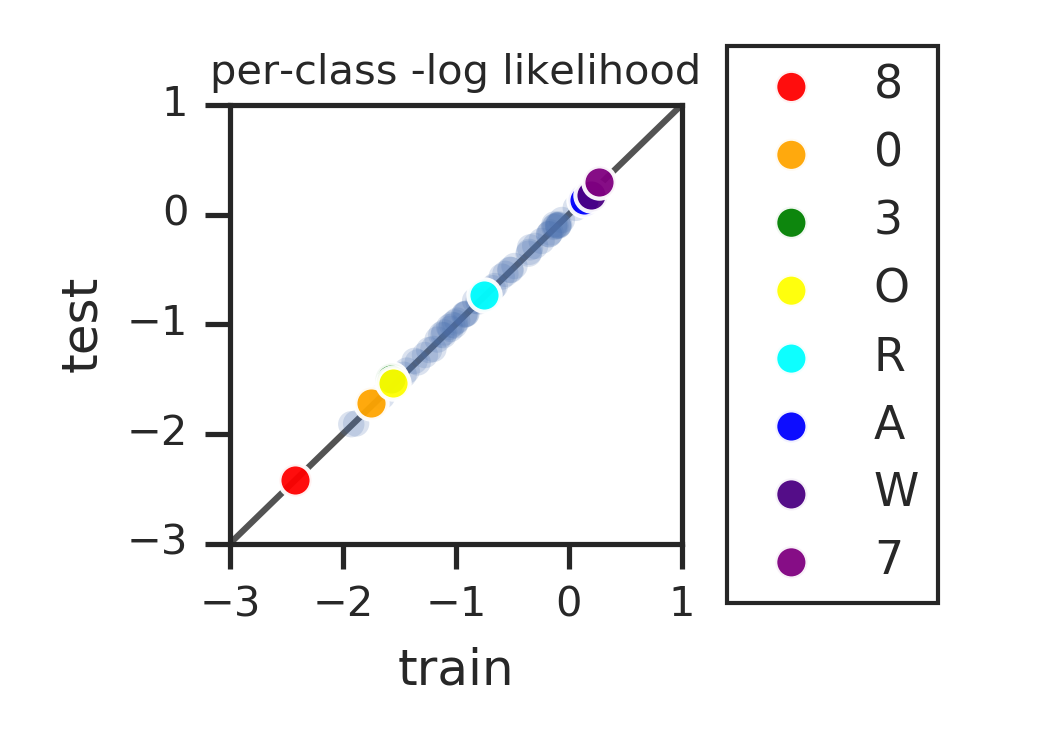}
        % \subcaption{\label{fig6:a}}%
    \end{minipage}%
    }%
    \hspace{0.05\textwidth}
    \subfloat[\label{fig6:b}]{
        \begin{minipage}[t]{0.416666\textwidth}
            \figsubtitle{Effect of Sequence Length on Decoding Quality}{\textwidth}
            \vspace{0.005in}\\
            \begin{minipage}[l]{\textwidth}
                \begin{minipage}[l]{0.6\textwidth}
                    \includegraphics[draft=false,width=\textwidth]{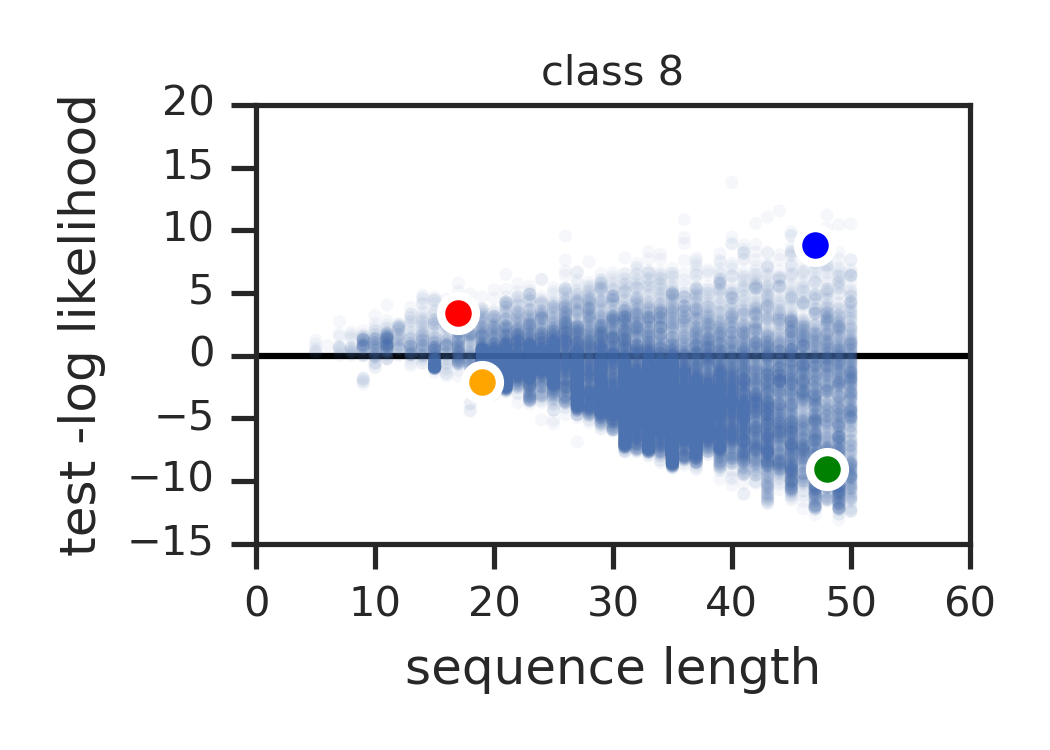}%
                \end{minipage}%
                \begin{minipage}[l]{0.2\textwidth}
                    \vspace{0.5\textwidth}
                    \includegraphics[draft=false,width=0.45\textwidth]{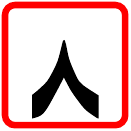}%
                    \includegraphics[draft=false,width=0.45\textwidth]{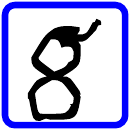}
                    \includegraphics[draft=false,width=0.45\textwidth]{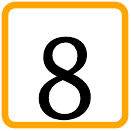}%
                    \includegraphics[draft=false,width=0.45\textwidth]{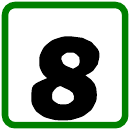}
                    \vspace{0.5\textwidth}
                \end{minipage}%
            \end{minipage}
            \begin{minipage}[l]{\textwidth}
                    \begin{minipage}[l]{0.6\textwidth}
                        \includegraphics[draft=false,width=\textwidth]{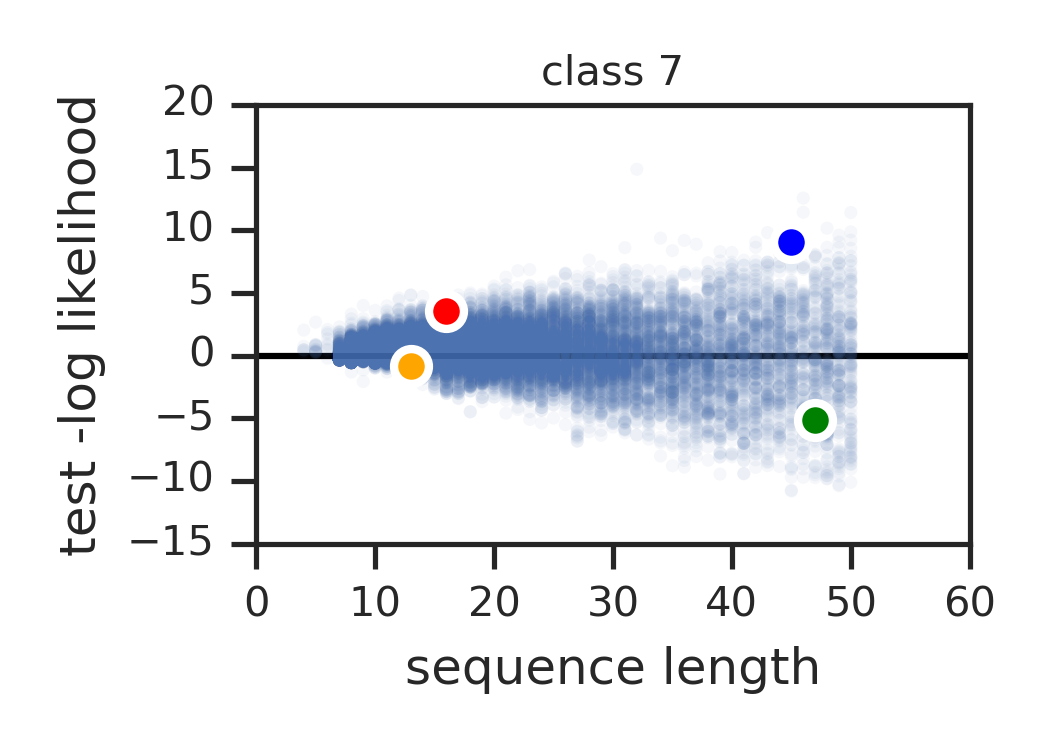}%
                        % \subcaption{\label{fig6:b}}%
                    \end{minipage}%
                \begin{minipage}[l]{0.2\textwidth}
                    \vspace{0.5\textwidth}
                    \includegraphics[draft=false,width=0.45\textwidth]{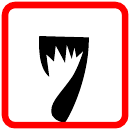}%
                    \includegraphics[draft=false,width=0.45\textwidth]{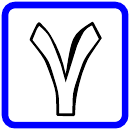}
                    \includegraphics[draft=false,width=0.45\textwidth]{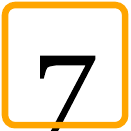}%
                    \includegraphics[draft=false,width=0.45\textwidth]{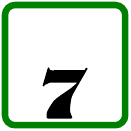}
                    \vspace{0.7\textwidth}
                    % \subcaption{\label{fig6:c}}%
                \end{minipage}%
            \end{minipage}%
        \end{minipage}%
    }
    \caption{\textbf{Quantifying the quality of the learned representations}. (a) Top: Negative log-likelihood for training and test datasets over $3$ epochs. Bottom: Negative log-likelihood for selected, individual classes within the dataset. (b, left) Test negative log-likelihood of all characters with label \classvalue{8} (top) and \classvalue{7} (bottom) as a function of the number of SVG commands. (b, right) Examples from \classvalue{8} and \classvalue{7}
    with few commands and high (red) or low loss (orange). Examples with many commands and high (blue) or low loss (green).}
    \label{fig6}
\end{figure*}

Almost all of the results presented have been assessed qualitatively. This is largely due to the fact that the quality of the results are assessed based on human judgements of aesthetics. In this section, we attempt to provide some quantitative assessment of the quality of the proposed model.

%So far we've characterized attributes of the learned representations in a qualitative way. Now, we will quantify their quality. 

Figure \ref{fig6:a} (top) shows the training dynamics of the model as measured by the overall training objective. Over the course of training $3$ epochs, we find that the model does indeed improve in terms of likelihood and plateaus in performance. Furthermore, the resulting model does not overfit on the training set in any significant manner as measured by the log-likelihood.

Figure \ref{fig6:a} (bottom) shows the mean negative log likelihoods for the examples in each class of the dataset. There is a small but systematic spread in average likelihood across classes. This is consistent with our qualitative results, where certain classes would consistently yield lower quality SVG decodings than others (e.g. \classvalue{7} in Figure \ref{fig:style-propagation}).

We can characterize the situations where the model performs best, and some possible causes for its improved performance. Figure \ref{fig6:b} shows the negative log likelihoods of examples from the test set of a given class, as a function of their sequence lengths. With longer sequences, the variance of log likelihoods increase. For the best performing class (\classvalue{8}, top) the loss values also trend downwards, whereas for the worst performing (\classvalue{7}, bottom), the trend stays relatively level. This means that the model had a harder time reliably learning characters especially with longer sequence lengths. 

Finally, in order to see what makes a given character hard or easy to learn, we examine test examples that achieved high and low loss, at different sequence lengths. Figure \ref{fig6:c}  reveals that for any class characters with high loss are generally highly stylized, regardless of their sequence lengths (red, blue), whereas easier to learn characters are more commonly found styles (yellow, green).

%-------------------------------------------------------------------------
\subsection{Limitations of working with a learned, stochastic, sequential representation}
\label{sec:limitations}

\begin{figure}
\begin{center}
    \begin{minipage}[t]{0.08\textwidth}
        \centering
        \figsubtitle{Common issues}{0.9\textwidth}
        \includegraphics[trim={0 0.1in 0 -0.15in},draft=false,width=1\textwidth]{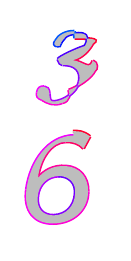}\\
        \vspace{0.27\textwidth}
    % \subcaption{\label{fig7:a}}
    \end{minipage}%
    %\begin{subfigure}[c]{0.4\textwidth}
        %\begin{center}
        %    \begin{subfigure}[r]{0.32\textwidth}
        %        \includegraphics[draft=false,height=0.6\textwidth]{fig/limitations/teal_x.png}\\
       %         \includegraphics[draft=false,width=0.6\textwidth]{fig/limitations/teal_7.png}
      %      \end{subfigure}%
      %      \begin{subfigure}[c]{0.38\textwidth}
                % \includegraphics[draft=false,width=\textwidth]{fig/limitations/precision_teal.pdf}
      %          % \includegraphics[draft=false,width=\textwidth]{fig/limitations/3_6_noteal.pdf}
     %           \includegraphics[draft=false,width=\textwidth]{fig/limitations/x_7.pdf}
    %            \includegraphics[draft=false,width=\textwidth]{fig/limitations/c_d_noteal.pdf}
    %        \end{subfigure}%
    %    \end{center}%
    %\subcaption{\label{fig7:b}}
    %\end{subfigure}
    \begin{minipage}[t]{0.36\textwidth}
        \vspace{-0.084\textwidth}
        \hspace{0.014\textwidth}\figsubtitle{Quantifying Model Confidence}{0.98\textwidth}
        \vspace{0.015\textwidth}\\
        \begin{minipage}[c]{0.4\textwidth}
            \includegraphics[trim={-0.3in 0 2in 0},draft=false,width=\textwidth]{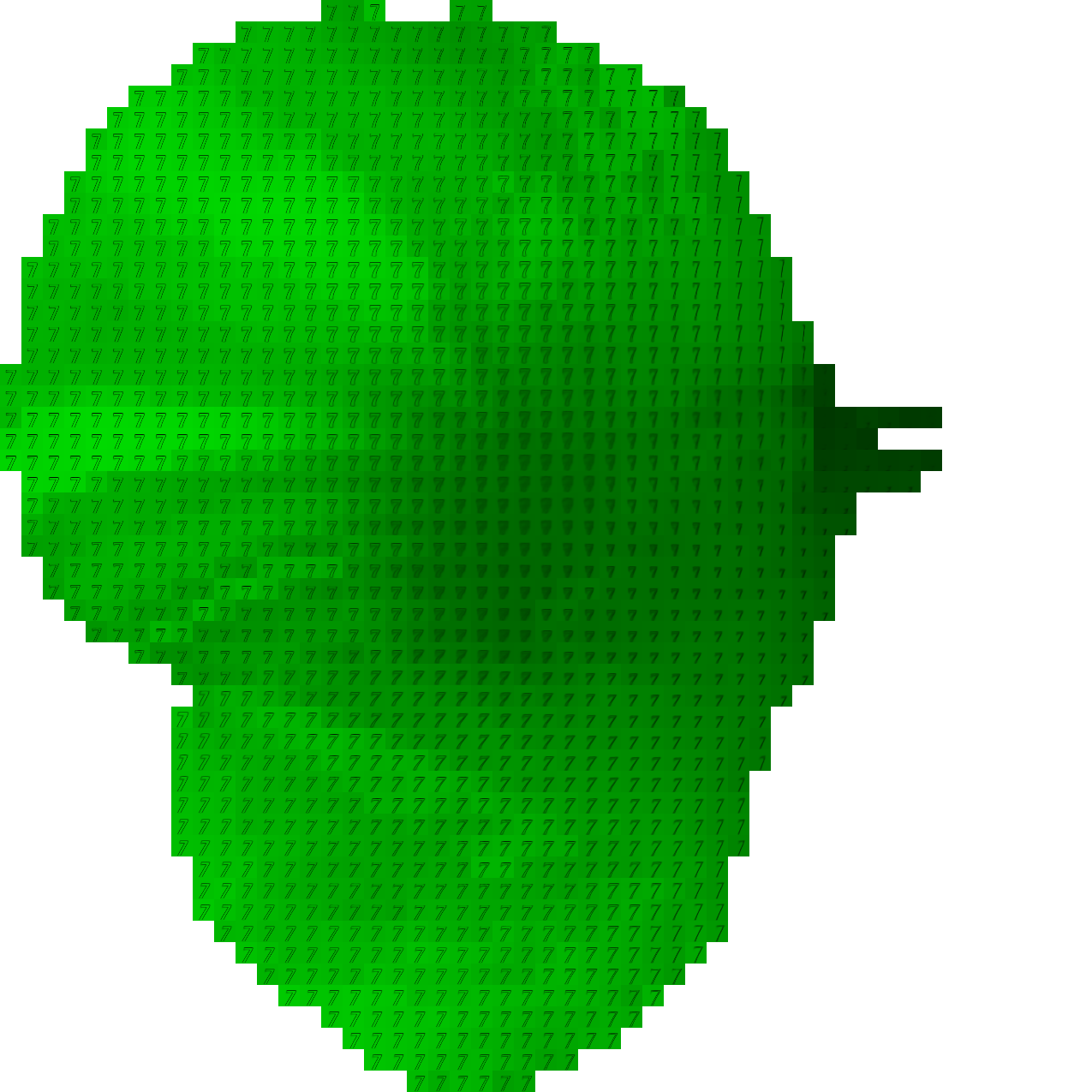}%
            \llap{\includegraphics[trim={-0.1618in 0 1.079215in 0},draft=false,width=1\textwidth]{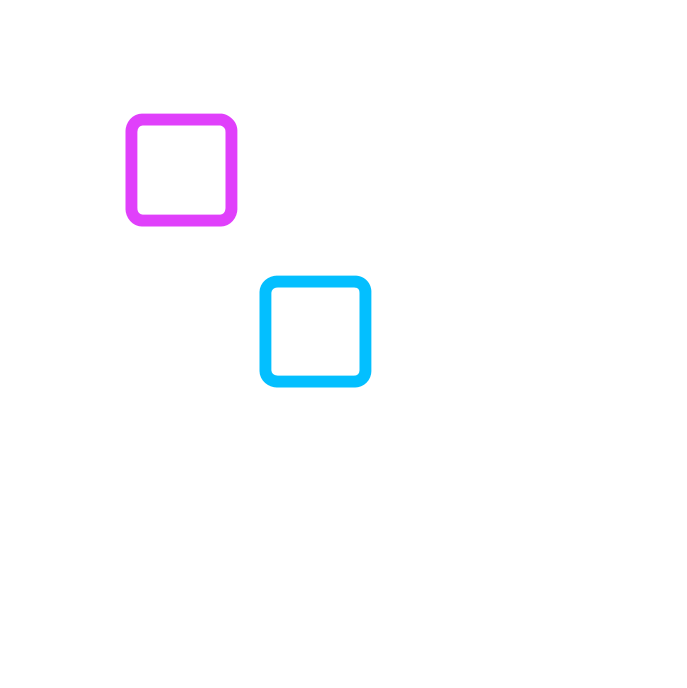}}%
            % \llap{\includegraphics[trim={-0.3in 0 2in 0},draft=false,width=1\textwidth]{fig/limitations/purple_yellow_boxes.png}}%
            % \llap{\includegraphics[trim={-0.3in 0 1in 0},draft=false,width=\textwidth]{fig/limitations/teal.png}}%
        \end{minipage}%
        \begin{minipage}[c]{0.6\textwidth}
            \centering
            \includegraphics[draft=false,width=0.5\textwidth]{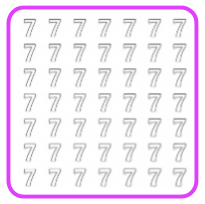}%
            \includegraphics[draft=false,width=0.5\textwidth]{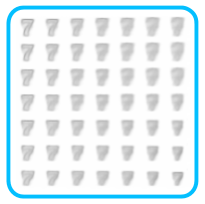}\\
            \vspace{0.013\textwidth}
            \includegraphics[draft=false,width=0.97\textwidth]{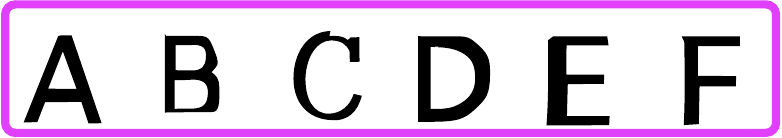}\\
            \vspace{0.02\textwidth}
            \includegraphics[draft=false,width=0.97\textwidth]{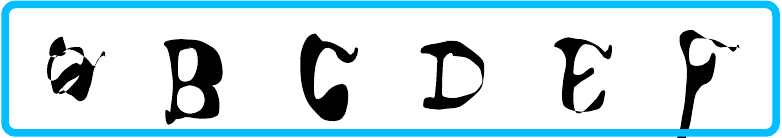}
        \end{minipage}%
    % \subcaption{\label{fig7:b}}
    \end{minipage}
\end{center}
\caption{\textbf{Limitations of proposed sequential, stochastic generative model.} Left: Low-likelihood samples may result in errors difficult to correct. Color indicates ordering of sequential sample (blue $\rightarrow$ red). Right: Regions of latent space with high variance result in noisy SVG decodings. Latent representation $z$ color coded based on variance: light (dark) green indicates low (high) variance. Visualization of rendered and SVG decoded samples (purple, blue).}
\label{fig:limitations}
\end{figure}

Given the systematic variability in the model performance across class label and sequence length, we next examine how specific features of the modeling choices may lead to these failures. An exhaustive set of examples of model failures is highlighted in in Figure \ref{fig:bad-random-fonts} of the Appendix. We discuss below two common modes for failure due to the sequential, stochastic nature of the generative model.

At each stochastic sampling step, the proposed model may select a low likelihood decision. Figure \ref{fig:limitations}a highlights how a mistake in drawing \classvalue{3} early on leads to a series of errors that the model was not able to error correct. Likewise, Figure \ref{fig7:a} shows disconnected start and end points in drawing  \classvalue{6} caused by the accumulation of errors over time steps. Both of these errors may be remedied through better training schedules that attempt to teach forms of error correction \cite{bengio2015scheduled}, but please see Discussion.

A second systematic limitation is reflected in the uncertainty captured within the model. Namely, the proposed architecture contains some notion of confidence in its own predictions as measured by the variance $\sigma^2$ in the VAE latent representation. We visualize the confidence by color-coding the UMAP representation for the latent style $z$ by $\sigma^2$ (Figure \ref{fig:limitations}b, right). Lighter green colors indicate high model confidence reflected by lower VAE variance.  Areas of high confidence show sharp outputs and decode higher quality SVGs (Figure \ref{fig:limitations}b, purple). Conversely, areas with lower confidence correspond to areas with higher label noise or more stylized characters. These regions of latent space decode lower quality SVGs (Figure \ref{fig:limitations}b, yellow). Addressing these systematic limitations is a modeling challenge for building next generation generative models for vector graphics (see Discussion).

\section{Discussion}

In the work we presented a generative model for vector graphics. This model has the benefit of providing a scale-invariant representation for imagery whose latent representation may be systematically manipulated and exploited to perform style propagation. We demonstrate these results on a large dataset of fonts and highlight the limitations of a sequential, stochastic model for capturing the statistical dependencies and richness of this dataset. Even in its present form, the current model may be employed as an assistive agent for helping humans design fonts in a more time-efficient manner \cite{carter2017using, rehling2001letter}. For example, a human may design a small set of characters and employ style propagation to synthesize the remaining set of characters (Figure \ref{fig:style-propagation}, \ref{fig:multi-style-propagation}).

An immediate question is how to build better-performing models for vector graphics. Immediate opportunities include new attention-based architectures \cite{vaswani2017attention} or potentially some form of adversarial training \cite{goodfellow2014generative}. Improving the model training to afford the opportunity for error correction may provide further gains \cite{bengio2015scheduled}.

A second direction is to employ this model architecture on other SVG vector graphics datasets. Examples include icons datasets \cite{clouatre2019figr} or human drawings \cite{sketchy2016, ha2017neural}. Such datasets reveal additional challenges above-and-beyond the fonts explored in this work as the SVG drawings encompass a larger amount of diversity and drawings containing larger numbers of strokes in a sequence. Additionally, employing color, brush stroke and other tools in illustration as a predicted features offers new and interesting directions for increasing the expressivity of the learned models.

\section*{Acknowledgements}
We'd like to thank the following people:
Diederik Kingma, Benjamin Caine, Trevor Gale, Sam Greydanus, Keren Gu, and Colin Raffel for discussions and feedback;
Monica Dinculescu and Shan Carter for insights from designers' perspective;
Ryan Sepassi for technical help using Tensor2Tensor;
Jason Schwarz for technical help using UMAP;
Joshua Morton for infrastructure help;
Yaroslav Bulatov, Yi Zhang, and Vincent Vanhoucke for help with the dataset;
and the Google Brain and the AI Residency teams.

{\small
\bibliographystyle{ieee}
\bibliography{egbib}
}

% Do this for final version.
% TODO: Acknowlwedge the people who built the glyph dataset? Compile a list of names for ths.
% Trevor Gale, Ben Caine, Keren Gu for discussions
% Colin Raffel, Ryan Sepassi, Dustin Tran for technical help with Tensor2Tensor
% idk if relevant by Sally, Leslie, Katie and others of the AI Residency team

% Joshua Morton for infrastructure help
% Monica Dinculescu and Shan Carter for feedback on designers' thought process when creating SVGs, and UX feedback
% Jason Schwarz for internal impl of UMAP and help using it
% Durk Kingma for review/discussion

% info from: https://g3doc.corp.google.com/experimental/users/vanhoucke/glyphs/README.md?cl=head#glyph200k
% Yaroslav Bulatov for mining the web for the font data
% Yi Zhang for helping put the raw data in placer and perform some initial data analysis
% Vincent Vanhoucke for some more analysis? or collection? or owning the data for a while?
\newpage

\clearpage
\appendix
\section{Appendix}

%------------------------------------------------------------------------------------------------
\subsection{Dataset details}
% \todo{right now there's a lot of duplication here with the "Data" section. Do we want to keep details here and remove them from there?}

% \todo{probably clarify language a bit. what is character/class vs glyph vs fontset, etc}

We collected a total of $14M$ font characters ($\mu=226,000$, $\sigma=22,625$ per class) in a common format (SFD), while retaining only characters whose unicode id corresponded to the classes \classvalue{0}-\classvalue{9}, \classvalue{a}-\classvalue{z}, \classvalue{A}-\classvalue{Z}. Filtering by unicode id is imperfect because many icons intentionally declare an id such that equivalent characters can be rendered in that font style
(e.g.: \begin{CJK*}{UTF8}{bsmi}七\end{CJK*} sometimes declares the unicode id normally reserved for \classvalue{7}).
% (e.g.: the character \begin{CJK*}{UTF8}{bsmi}七\end{CJK*} in some fonts sometimes declares the unicode id normally reserved for \classvalue{7}).

% Dataset statistics: $14$M examples, average of $226,000$ and standard deviation of $22,625$ for each of the 62 classes (\classvalue{0}-\classvalue{9}, \classvalue{a}-\classvalue{z}, \classvalue{A}-\classvalue{Z}).

% The glyphs were collected by scraping the web for fontsets in a common font format (SFD). We filtered out glyphs whose unicode id does not correspond to the classes we want to model. This is imperfect because many icons have incorrect unicode ids or intentionally declare a unicode id so that equivalent characters can be rendered in that font style (e.g.: fontsets that use the unicode id normally reserved for \classvalue{7} for the glyph \begin{CJK*}{UTF8}{bsmi}七\end{CJK*}).

We then convert the SFD icons into SVG. The SVG format can be composed of many elements (\svg{square}, \svg{circle}, etc). The most expressive of these is the \svg{path} element whose main attribute is a sequence of commands, each requiring a varying number of arguments (\svg{lineTo}: 1 argument, \svg{cubicBezierCurve}: 3 arguments, etc.). An SVG can contain multiple elements but we found that SFD fonts can be modelled with a single \svg{path} and a subset of its commands (\svg{moveTo}, \svg{lineTo}, \svg{cubicBezierCurve}, \svg{EOS}). This motivates our method to model SVGs as a single sequence of commands.

% We then convert the SFD glyphs into SVG format. The SVG format can be composed of many hierarchical elements, with strong constructs (e.g.: \svg{square}, \svg{circle}). The most expressive of these is the \svg{path} element, which defines a sequence of "commands" (e.g.: \svg{lineTo}, \svg{cubicBezierCurve}, etc). Each of these commands require a varying number of arguments (e.g.: \svg{lineTo} needs a final x,y position, \svg{cubicBezierCurve} needs two control points and a final x,y position). An SVG can also contain multiple "path" elements. However, we found that the vast majority of icons found online can be modelled with a single "path" element. Further, all the fonts can be converted from SFD to SVG using a single "path" element and a subset of its commands (\svg{moveTo}, \svg{lineTo}, \svg{cubicBezierCurve}, \svg{EOS}). This motivates our method to model SVGs as a single sequence of commands.

% The distribution of the number of commands in a given path has a long tail.
In order to aid learning, we filter out characters with over 50 commands.
% , which still leaves plenty of room for highly stylized icons.
We also found it crucial to use relative positioning in the arguments of each command. Additionally, we re-scale the arguments of all icons to ensure that most real-values in the dataset will lie in similar ranges. This process preserves size differences between icons. Finally, we standardize the command ordering within a path such that each shape begins and ends at its top-most point and the curves always start by going clockwise. We found that setting this prior was important to remove any ambiguity regarding where the SVG decoder should start drawing from and which direction (information which the image encoder would not be able to provide).

% \svg{path} allows the use of relative or absolute positioning for the arguments given to these commands. We found the use of relative positioning crucial for learning. Further, the distribution of the number of commands in a given path has a very long tail. In order to aid learning, we filter out characters with over 50 commands, which still leaves plenty of room for highly stylized icons.

% The SVG \svg{viewbox} element determine's the region of euclidean space that is visible when the SVG is rendered. For instance, if the viewbox is the $24$x$24$ square region, then curves within this region will be visible and anything outside will be cut out. The size of this region is not standardized across fronts so when converting the SFD fonts to SVG, we re-scale all commands arguments by the ratio of the font's viewbox with respect to a fixed $24$x$24$ viewbox. This way, we guarantee that most real-values in the dataset will lie in similar ranges, while also preserving size differences between glyphs (e.g.: an icon that took most of the space in its viewbox will also take up most of the space in the standardized viewbox, and vice versa).

% We also standardize the command ordering within a path such that each shape begins and ends at its top-most point and the curves always start by going clockwise. We found that setting this prior was important to remove any ambiguity regarding where the SVG decoder should start drawing from and which direction (information which the image encoder would not be able to provide).

When rendering the SVGs as $64$x$64$ pixel images, we pick a render region that captures tails and descenders that go under the character's baseline. Because relative size differences between fonts are still present, this process is also imperfect: zoom out too little and many characters will still go outside the rendered image, zoom out too much and most characters will be too small for important style differences to be salient.

% Then we render the SVGs as a $64$x$64$ pixel image. We use a viewbox that is slightly larger than the normalized one mentioned above to capture tails and descenders that go under the character's baseline. Because relative size differences between fonts are still present, this process is also imperfect: zoom out too little and many glyphs will still go outside the rendered image, zoom out too much and most glyphs will be too small for important style differences to be salient.

% Then we render the SVGs as a $64$x$64$ pixel image. We found that most fonts used the bottom edge of their viewboxes as the font's baseline. Many characters (e.g.: \classvalue{g}) have elements that go under the baseline, so we render using a viewbox that is slightly larger than the normalized one mentioned above. Because relative size differences between fonts are still present, this process is also imperfect: zoom out too little and many glyphs will still go outside the rendered image, zoom out too much and most glyphs will be too small for important style differences to be salient.

Lastly, we convert the SVG path into a vector format suitable for training a neural network model: each character is represented by a sequence of commands, each consisting of tuples with: 1) a one-hot encoding of command type (\svg{moveTo}, \svg{lineTo}, etc.) and 2) a normalized representation of the command's arguments (e.g.: x, y positions). Note that for this dataset we only use 4 commands (including EOS), but this representation can easily be extended for any SVG icon that use any or all of the commands in the SVG path language.

%------------------------------------------------------------------------------------------------------
\subsection{Details of network architecture}

% \todo{consider making this much shorter}
Our model is composed of two separate substructures: a convolutional variational autoencoder and an auto-regressive SVG decoder. The model was implemented and trained with Tensor2Tensor \cite{vaswani2018tensor2tensor}.

The image encoder is composed of a sequence of blocks, each composed of a convolutional layer, conditional instance normalization (CIN) \cite{dumoulin2017learned, perez2018film}, and a ReLU activation. Its output is a $z$ representation of the input image. 
At training time, $z$ is sampled using the reparameterization trick \cite{kingma2013auto, rezende2014stochastic}. At test time, we simply use $z = \mu$.
The image decoder is an approximate mirror image of the encoder, with transposed convolutions in place of the convolutions. All convolutional-type layers have \texttt{SAME} padding. CIN layers were conditioned on the icon's class.
% (\classvalue{0}-\classvalue{9}, \classvalue{a}-\classvalue{z}, \classvalue{A}-\classvalue{Z}). 
% The image encoder's final layer outputs a $z$ representation of the input image, in the form of a mean $\mu$ and the log of the standard deviation $log(\sigma)$. At training time, we use the reparameterization trick \cite{kingma2013auto, rezende2014stochastic} to sample $z$ from these statistics, which is then fed to the image decoder. At test time, we simply use $z = \mu$.

\begin{table}[h]
\begin{center}
\begin{tabular}{ |c|c|c|c|c| } 
 \hline
 Operations & Kernel, Stride & Output Dim \\
 \hline
 Conv-CIN-ReLU & 5, 1 & 64x64x32 \\ 
 Conv-CIN-ReLU & 5, 1 & 32x32x32 \\ 
 Conv-CIN-ReLU & 5, 1 & 32x32x64 \\ 
 Conv-CIN-ReLU & 5, 2 & 16x16x64 \\
 Conv-CIN-ReLU & 4, 2 & 8x8x64 \\
 Conv-CIN-ReLU & 4, 2 & 4x4x64 \\
 Flatten-Dense & - & 64 \\
 \hline
\end{tabular}
%\begin{tabular}{ |c|c| } 
% \hline
% Total parameters & $416,672$ \\
% \hline
%\end{tabular}
\end{center}
\label{table:image-encoder}
\caption{Architecture of convolutional image encoder containing $416,672$ parameters.}
\end{table}

% The image encoder's final layer outputs a $z$ representation of the input image, in the form of a mean $\mu$ and the log of the standard deviation $log(\sigma)$. At training time, we use the reparameterization trick \cite{kingma2013auto, rezende2014stochastic} to sample $z$ from these statistics, which is then fed to the image decoder. At test time, we simply use $z = \mu$.

\begin{table}[h]
\begin{center}
%Image decoder
\begin{tabular}{ |c|c|c|c|c| } 
 \hline
 Operations & Kernel, Stride & Output Dim \\
 \hline
 Dense-Reshape & - & 4x4x64 \\ 
 ConvT-CIN-ReLU & 4, 2 & 8x8x64 \\ 
 ConvT-CIN-ReLU & 4, 2 & 16x16x64 \\
 ConvT-CIN-ReLU & 5, 1 & 16x16x64 \\
 ConvT-CIN-ReLU & 5, 2 & 32x32x64 \\
 ConvT-CIN-ReLU & 5, 1 & 32x32x32 \\
 ConvT-CIN-ReLU & 5, 2 & 64x64x32 \\
 ConvT-CIN-ReLU & 5, 1 & 64x64x32 \\
 Conv-Sigmoid & 5, 1 & 64x64x1 \\
 \hline
\end{tabular}
%\begin{tabular}{ |c|c| } 
% \hline
% Total parameters & $516,865$ \\
% \hline
%\end{tabular}
\end{center}
\label{table:image-decoder}
\caption{Architecture of convolutional image decoder containing $516,865$ parameters.}
\end{table}

The SVG decoder consists of 4 stacked LSTMs cells with hidden dimension $1024$, trained with feed-forward dropout \cite{srivastava2014dropout}, as well as recurrent dropout \cite{zaremba2014recurrent,semeniuta2016recurrent} at $70\%$ keep-probability. The decoder's topmost layer consists of a Mixture Density Network (MDN) \cite{bishop1994mixture, graves2013generating}. It's hidden state is initialized by conditioning on $z$. At each time-step, the LSTM receives as input the previous time-step's sampled MDN output, the character's class and the $z$ representation. The total number of parameters is $34,875,272$.

% We use 4 stacked LSTM cells, with hidden dimension $1024$. Each cell is a basic LSTM cell, with feedforward and recurrent dropout, both with keep-probability $70\%$. The hidden state of each layer is conditioned on $z$ through a dense layer that outputs a tensor with the appropriate dimensionality. At each step, a Dense-Tanh layer conditioned on the concatenation of the previous step's output, $z$, and the glyph's class, outputs the input to the bottom-most cell with appropriate dimensionality. The total number of parameters is $34,875,272$.

%------------------------------------------------------------------------------------------------
\subsection{Training details}

The optimization objective of the image VAE is the log-likelihood reconstruction loss and the KL loss applied to $z$ with KL-beta $4.68$. We use Kingma et al's \cite{kingma2016improved} trick with $4.8$ free bits ($0.15$ per dimension of $z$). The model is trained with batch size $64$.

The SVG decoder's loss is composed of a softmax cross-entropy loss between the one-hot command-type encoding, added to the MDN loss applied to the real-valued arguments. We found it useful to scale the softmax cross-entropy loss by $10$ when training with this mixed loss. We trained this model with teacher forcing and used batch size $128$.

All models are initialized with the method proposed by He et al. (2015) \cite{he2015delving} and trained with the Adam optimizer with $\epsilon = 10^{-6}$ \cite{kingma2014adam}.

%------------------------------------------------------------------------------------------------
\subsection{Visualization details}

% \todo{UMAP parameters (1M examples, hyper parameters used, exactly how we discretized the grid, etc.)}

The dimensionality reduction algorithm used for visualizing the latent space is UMAP \cite{mcinnes2018umap}. We fit the activations $z$ of $1$M examples from the dataset into $2$ UMAP components, using the cosine similarity metric, with a minimum distance of $0.5$, and $50$ nearest neighbors. After fitting, we discretize the 2D space
% by dividing its longest axis (x or y)
into $50$ discrete buckets
% , and dividing the remaining axis into buckets with the same length. The $32$-dimensional $z$s which lied in the same 2D grid were averaged and used as conditioning for the image decoder.
and decode the average $z$ in each grid cell with the image decoder.

%------------------------------------------------------------------------------------------------
\subsection{Samples from generated font sets}
\begin{figure*}[t]
    \begin{center}
    \includegraphics[draft=false,height=1.35cm]{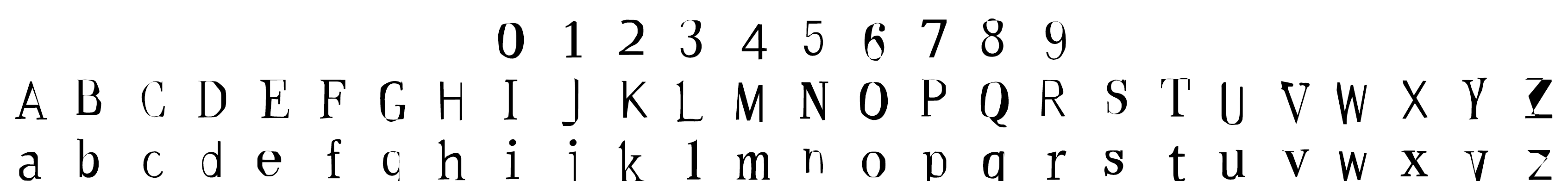}
    \includegraphics[draft=false,height=1.35cm]{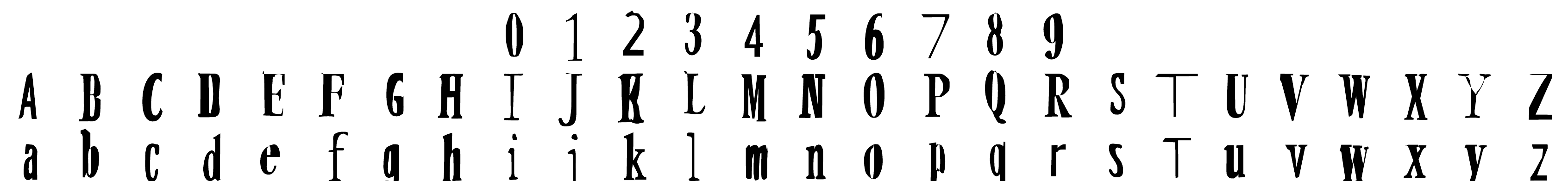}
    \includegraphics[draft=false,height=1.35cm]{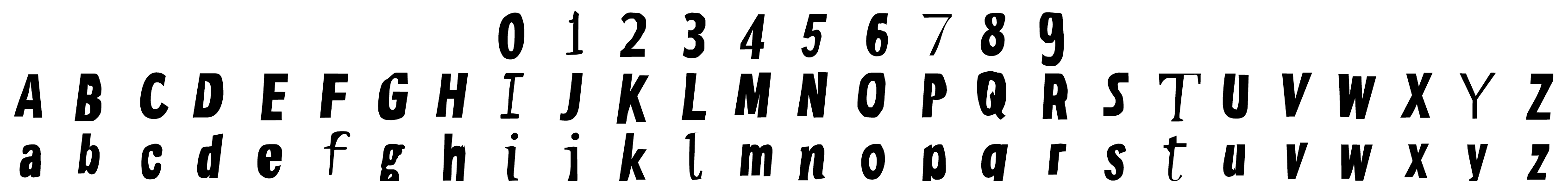}    
    \includegraphics[draft=false,height=1.35cm]{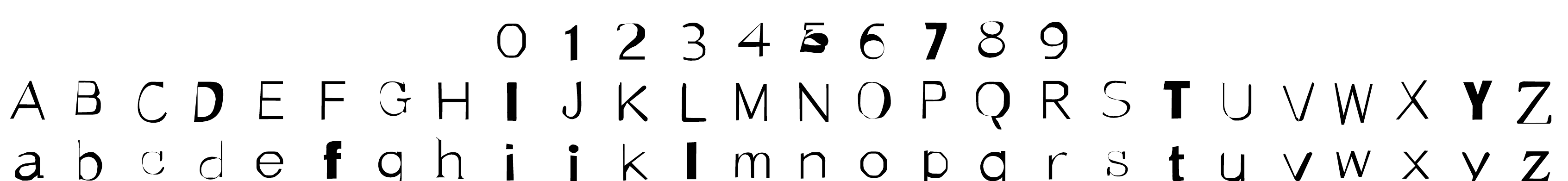}
    \includegraphics[draft=false,height=1.35cm]{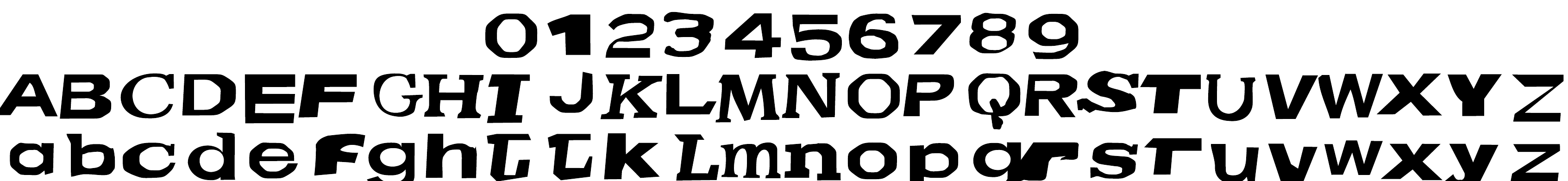}
    \includegraphics[draft=false,height=1.35cm]{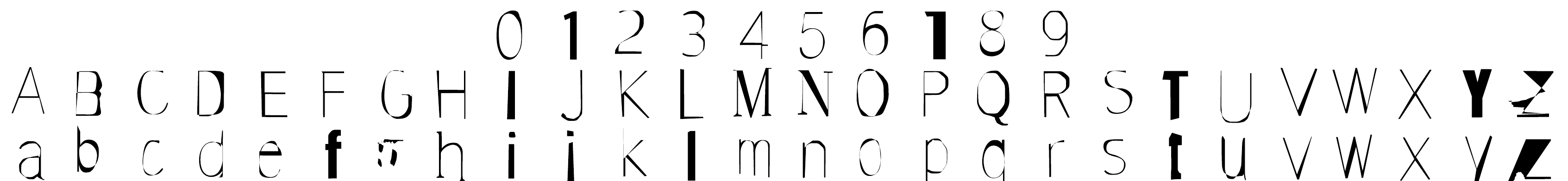}
    \includegraphics[draft=false,height=1.35cm]{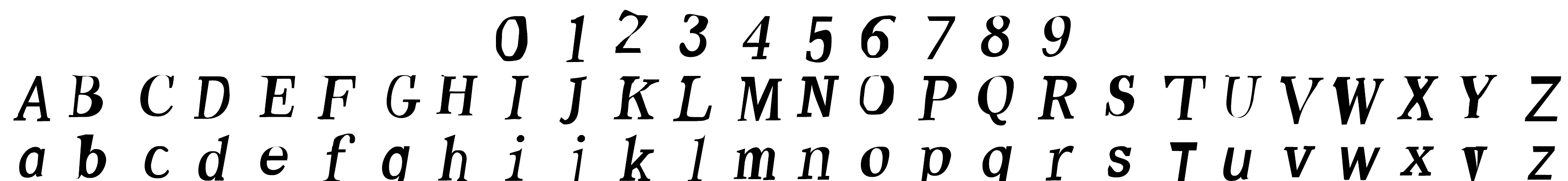}
    \includegraphics[draft=false,height=1.35cm]{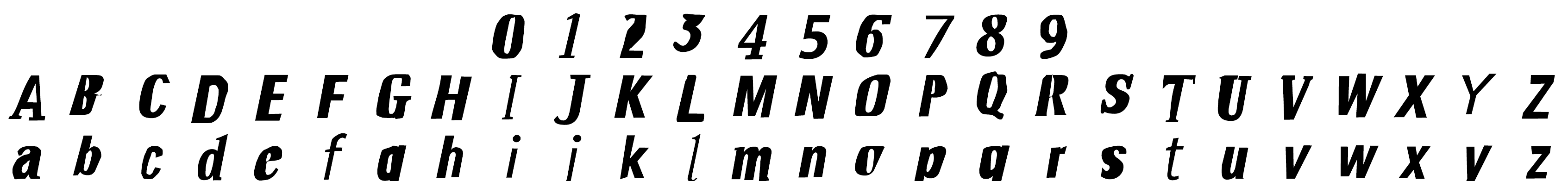}
    \includegraphics[draft=false,height=1.35cm]{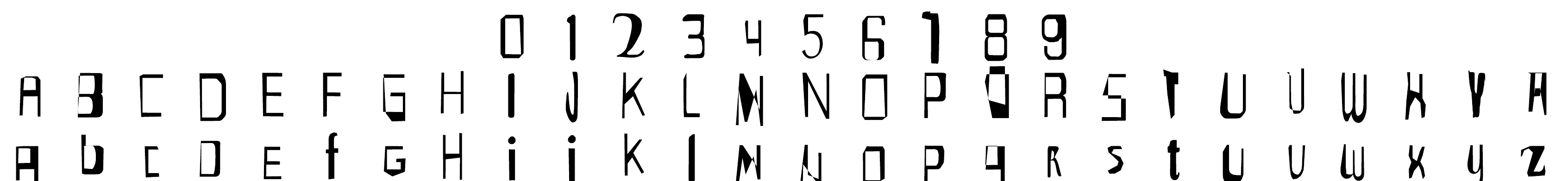}
    \end{center}
   \caption{\textbf{More examples of randomly generated fonts}. Details follow figure \ref{fig:random-fonts}.}
    \label{fig:more-random-fonts}
\end{figure*}

\begin{figure*}
    \begin{center}
    % Todo: separate the bad ones from here into a new figure and discuss why the model makes those mistakes.
    \includegraphics[draft=false,height=1.35cm]{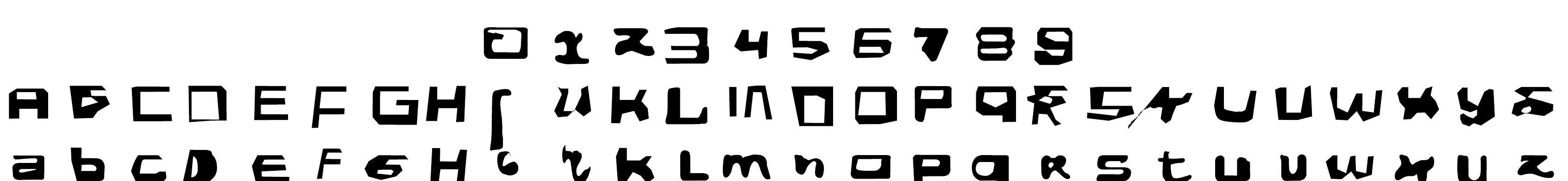}
    \includegraphics[draft=false,height=1.35cm]{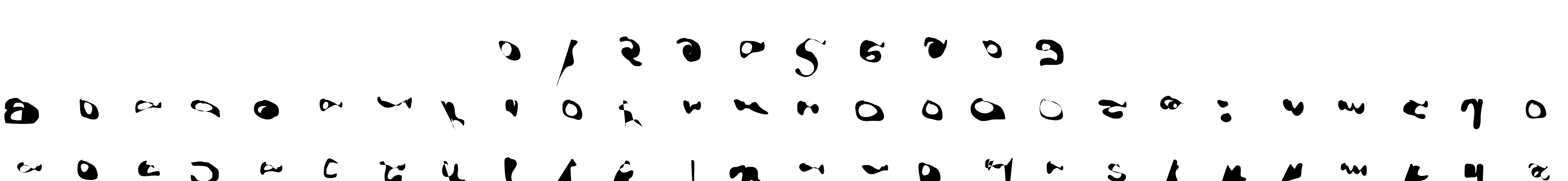}
    \includegraphics[draft=false,height=1.35cm]{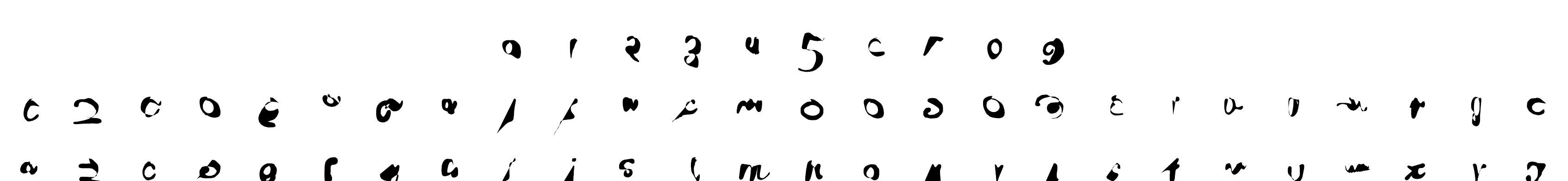}
    \end{center}
   \caption{\textbf{Examples of poorly generated fonts}. Details follow figure \ref{fig:random-fonts}. Highly stylized characters clustered in a high-variance region of the latent space $z$. Samples from this region generate poor quality SVG fonts. See Section \ref{sec:limitations} for details.}
    \label{fig:bad-random-fonts}
\end{figure*}

Figures \ref{fig:svg}, \ref{fig:smooth-latent}, \ref{fig:style-propagation}, \ref{fig:style-analogies} and \ref{fig:limitations} contained selected examples to highlight the successes and failures of the model as well as demonstrate various results. To provide the reader with a more complete understanding of the model performance, below we provide additional samples from the model highlighting successes (Figure \ref{fig:more-random-fonts}) and failures (Figure \ref{fig:bad-random-fonts}). As before, results shown are selected best out of 10 samples.

\end{document}